\documentclass{article}
\usepackage[margin=1in]{geometry}
\usepackage{graphicx} 
\usepackage{xcolor}
\usepackage{amsmath, amsthm, amssymb, amsfonts}
\usepackage[colorlinks=true,linkcolor=blue,citecolor=blue]{hyperref}
\usepackage{bm}
\usepackage{subcaption}
\usepackage{natbib}
\usepackage{booktabs}
\usepackage{multirow}
\usepackage{multicol}

\title{On the accuracy of implicit neural representations for cardiovascular anatomies and hemodynamic fields}
\author{Jubilee Lee, Daniele E. Schiavazzi}
\date{Department of Applied and Computational Mathematics and Statistics\\
\vspace{5pt}
University of Notre Dame
}

\begin{document}

\maketitle

\begin{abstract}
\noindent Implicit neural representations (INRs, also known as \emph{neural fields}) have recently emerged as a powerful framework for knowledge representation, synthesis, and compression.
By encoding fields as continuous functions within the weights and biases of deep neural networks—rather than relying on voxel- or mesh-based structured or unstructured representations—INRs offer both resolution independence and high memory efficiency.
However, their accuracy in domain-specific applications remains insufficiently understood. In this work, we assess the performance of state-of-the-art INRs for compressing hemodynamic fields derived from numerical simulations and for representing cardiovascular anatomies via signed distance functions.
We investigate several strategies to mitigate spectral bias, including specialized activation functions, both fixed and trainable positional encoding, and linear combinations of nonlinear kernels.
On realistic, space- and time-varying hemodynamic fields in the thoracic aorta, INRs achieved remarkable compression ratios of up to approximately 230, with maximum absolute errors of 1 mmHg for pressure and 5 to 10 cm/s for velocity, without extensive hyperparameter tuning.
Across 48 thoracic aortic anatomies, the average and maximum absolute anatomical discrepancies were below 0.5 mm and 1.6 mm, respectively. Overall, the SIREN, MFN-Gabor, and MHE architectures demonstrated the best performance. Source code and data is available at \url{https://github.com/desResLab/nrf}.
\end{abstract}



\section{Introduction}\label{sec:intro}

The growing availability of computational power, combined with recent advances in accelerated computing~\cite{silvano2025survey}, has substantially reduced the effort required to generate realistic multiphysics simulations of complex processes across many fields.
As a recurring theme throughout this paper, cardiovascular modeling has evolved rapidly from the conceptual foundations laid by Harvey and Frank~\cite{harvey2014works,sutera1993history} to full-fledged three-dimensional simulations with hundreds of millions of degrees of freedom~\cite{verzicco2022electro}. These simulations capture the coupled interactions among fluid~\cite{bazilevs2007variational}, structure~\cite{bazilevs2013computational}, and electrophysiology~\cite{viola2020fluid,regazzoni2022cardiac}, as well as multiscale temporal phenomena such as growth and remodeling~\cite{pfaller2024fsge}.

However, the size of the result fields generated by such simulators is also increasing rapidly, posing significant challenges in managing the large volumes of data required for downstream tasks such as analysis, visualization, and information extraction, as well as for outer-loop tasks including optimization~\cite{marsden2014optimization}, uncertainty quantification~\cite{schiavazzi2016uncertainty,tran2017automated,fleeter2020multilevel}, inference~\cite{MENON2025108951,tong2025invaert,choi2025performance}, and the development of digital twins for decision-making~\cite{sel2024building,lee2024probabilistic,tudor2025scoping}.

Recent advances in deep neural networks have considerably strengthened both their theoretical foundations and practical training methodologies, including \emph{existence} results establishing their nature as \emph{universal approximators}~\cite{kidger2020universal}, efficient optimization through gradient descent~\cite{rumelhart1986learning}, and modern software frameworks that facilitate training across a wide range of architectures~\cite{paszke2019pytorch}.
These developments have led to the emergence of implicit neural representations (INRs, or \emph{neural fields}), which can encode complex space–time fields within the weights and biases of multilayer perceptrons. 
Other emerging techniques for fast rendering of three-dimensional fields -- such as Gaussian splatting~\cite{kerbl20233d} -- have also gained attention; however, this paper focuses exclusively on INRs.

Neural network–based approaches, however, are known to suffer from the phenomenon of \emph{spectral bias}, referring to the tendency of multilayer perceptrons to preferentially learn low-frequency components of the data during training~\cite{rahaman2019spectral,xu2019frequency}.
Efforts to mitigate this issue include (1) the design of activation functions capable of capturing a broader range of frequencies, (2) the development of trainable input encoding, (3) adjustments to training dynamics through normalization and gradient modulation, (4) the adoption of specialized training strategies, (5) the construction of architectures specifically tailored to multiscale PDEs, and others.
It is worth noting, however, that spectral bias is not always undesirable. In certain contexts, it can even be advantageous -- for instance, by suppressing high-frequency error components in hybrid data-driven and traditional preconditioners for large systems of linear equations~\cite{kopanivcakova2025deeponet}.

A large body of work has addressed spectral bias through carefully designed neural activation functions.
SIREN~\cite{sitzmann2020implicit} employs sine activations to enhance the representation of fine details, while FINER introduces variable-periodic activation functions~\cite{liu2024finer}. A harmonic analysis of sinusoidal networks led to TUNER~\cite{novello2025tuning}, in which activation frequencies are initialized through spectral sampling and weight clamping is applied to bound the maximum representable frequency. Similarly, STAF~\cite{morsali2025staf} achieves a trainable activation frequency spectrum, and WIRE~\cite{saragadam2023wire} employs Gabor wavelet activations for their optimal space–frequency localization properties. Another related approach, the \emph{multiplicative filter network}~\cite{fathony2021multiplicative}, combines the efficiency of neural architectures with the expressive power of traditional spectral and multiresolution representations, by composing linear combinations of nonlinear Fourier- or Gabor-like filters. These architectures are discussed further in Section~\ref{sec:mfn}.

Other research directions have focused on addressing spectral bias through positional encoding. Fixed-frequency and random (Gaussian) frequency encoding were introduced in~\cite{tancik2020fourier} and~\cite{mildenhall2021nerf}, respectively.
State-of-the-art methods include multiresolution hash encoding (MHE~\cite{muller2021real,muller2022instant} further discussed in Section~\ref{sec:mhe}), which has been demonstrated across applications such as large-scale image reconstruction, neural radial fields (NeRF), signed-distance function representations of geometries, and volume rendering.
Another active line of work has explored training strategies and gradient manipulation. Batch normalization (BN~\cite{ioffe2015batch}) effectively shifts the spectrum of the neural tangent kernel toward higher values, enabling networks to capture high-frequency components~\cite{cai2024batch}. High-frequency learning has also been achieved through \emph{stacking} and incremental training of shallow networks~\cite{fang2024addressing}, while extensions replacing Fourier features with quantum circuits have been proposed in~\cite{zhao2024quantum}.

Spectral bias has also been identified as a key limitation in scientific deep learning, particularly in modeling fine-scale variations such as boundary layers or multiphase flows. High-Frequency Scaling (HFS) for neural operators~\cite{khodakarami2025mitigating} amplifies high-frequency components by scaling latent Fourier representations in the decoder, improving predictions for multiphase and chaotic systems. Since multiscale PDEs exacerbate spectral bias, HANO~\cite{liu2024mitigating} introduces a hierarchical attention mechanism that enables cross-scale interactions and employs an empirical $H^1$ loss to emphasize gradient accuracy and better capture high-frequency behavior. PhaseDNN~\cite{cai2020phase} decomposes problems into low-frequency sub-problems, training separate networks on phase-shifted data and combining their outputs. The Incremental Fourier Neural Operator (iFNO)~\cite{george2022incremental} progressively increases frequency modes and spatial resolution during training to prevent premature convergence to low frequencies.
Physics-Informed Residual Adaptive Networks (PirateNets)~\cite{wang2024piratenets} mitigate performance degradation in large, deep physics-informed architectures by introducing adaptive residual connections and initialization schemes that encode inductive biases from the underlying PDE system. An application of PirateNets to turbulent flow reconstruction is presented in~\cite{wang2025simulating}.
From a computer-graphics and visualization perspective, recent work has focused on accelerating neural field generation through hypernetworks~\cite{wu2023hyperinr}, distributed implementations~\cite{wu2024distributed}, and cache-accelerated representations~\cite{zavorotny2025cluster}.

A particularly interesting unified framework treats neural field representations (i.e., \emph{functions}) directly as data -- so-called \emph{functa}~\cite{dupont2022data} -- enabling their use in complex inference, generative modeling, and classification tasks in place of grid-based data. Extensions to this concept improve scaling behavior~\cite{bauer2023spatial} and extend its application to PDEs via global Fourier modulation (GFM), which re-parameterizes each layer using Fourier kernels~\cite{jo2025pdefuncta}.

While several studies have applied physics-informed loss augmentation (PINNs~\cite{karniadakis2021physics}) and deep operator networks~\cite{lu2021learning} to cardiovascular flow problems~\cite{zhang2023physics,dermul2024reconstruction}, relatively few have explicitly investigated spectral bias mitigation within the context of implicit neural representations. The contribution in~\cite{saitta2024implicit} employed INRs with SIREN networks for unsupervised super-resolution and denoising of four-dimensional MRI flow fields. More recent studies have used INRs to characterize motion and deformation in the vascular lumen and cardiac wall~\cite{lowes2024implicit,bell2025implicit}, or for three-dimensional reconstruction of the left-ventricular anatomy~\cite{sander2023reconstruction}. In the latter, an auto-decoder MLP maps three-dimensional coordinates to an occupancy or signed-distance function, learning high-resolution left ventricular geometries from CT data. INRs and NeRFs have also been used to recover the three-dimensional geometry of coronary vessels from two X-ray angiograms~\cite{wang2024neca,maas2025nerf}, while Kong et al.~\cite{kong2024sdf4chd} proposed a generative implicit shape model using signed-distance fields (SDFs) to represent whole-heart anatomy across various congenital defects.
Finally, Zhang et al.~\cite{zhang2025shape} introduced a non-intrusive approach for building surrogate models that approximate the solutions of parameterized PDEs while accounting for geometric variations in the computational domain.

Despite substantial progress in both representation quality and training efficiency of neural fields~\cite{muller2022instant}, most evaluations have focused on two-dimensional images, geometric models, or NeRF reconstructions. 
Thus, significant gaps remain in understanding how different neural network architectures perform when representing physically relevant fields in scientific applications.
In this paper, we focus on result fields from steady-state and pulsatile hemodynamic simulations computed using stabilized and variational multiscale finite elements within the SimVascular solver platform~\cite{updegrove2017simvascular}. Our contributions are as follows:
\begin{itemize}\itemsep 0pt
\item We provide a comprehensive evaluation of state-of-the-art implicit neural representations for cardiovascular pressure and velocity fields across both patient-agnostic and patient-specific models.
\item We quantify the accuracy of neural field–based geometric representations for both idealized and patient-specific vascular anatomies and analyze the factors influencing their performance.
\end{itemize}

Source code and data for reproducing the results presented in the following sections are available at \url{https://github.com/desResLab/nrf}. 
Section~\ref{sec:methods} presents the theoretical background and implementation details of nine neural field architectures. Section~\ref{sec:results} reports their performance in reconstructing fields from two to four dimensions (Sections~\ref{sec:xt_pipe}–\ref{sec:xyzt_aorta}), including detailed results for a thoracic aorta model comprising approximately three million elements and 180 time steps.
Sections~\ref{sec:sdf} and~\ref{sec:sdf_zoo} extend this analysis to signed-distance function representations for one idealized model, one pulmonary anatomy, and one thoracic aortic anatomy, followed by 48 additional patient-specific thoracic aortas from the Vascular Model Repository~\cite{wilson2013vascular}.
We conclude with a discussion and final remarks in Section~\ref{sec:discussion_conclusions}.

\section{Methods}\label{sec:methods}

\subsection{INR Architectures}\label{sec:inr_architectures}

In this section we discuss several network architectures for implicit neural representations proposed in the literature. Starting from the widely adopted Multilayer Perceptron (MLP) baseline, we discuss a-priori selected frequency-based positional encoding~\cite{tancik2020fourier}, SIREN~\cite{sitzmann2020implicit}, multiplicative filter networks~\cite{fathony2021multiplicative}, and conclude with multiresolution hash encoding (MHE~\cite{muller2022instant}).

\subsubsection{MLP and frequency-based non-trainable positional encoding}

MLP are feedforward neural networks with fully connected input, hidden, and output layers employing nonlinear activation functions (e.g., sigmoid, tanh, ReLU). 
This technology has been extensively applied for classification, regression, surrogate modeling, and time series forecast across numerous applications in science and engineering, and remains a core architecture in neural network research (see, e.g.,~\cite{goodfellow2016deep}). 
Combining MLP with positional encoding (PE~\cite{tancik2020fourier}) has been proven effective in mitigating spectral bias.
We consider two approaches for PE, the first using randomly selected frequencies (referred to as \emph{Gaussian} positional encoding in~\cite{tancik2020fourier}), and a second approach where such frequencies are defined a-priori~\cite{mildenhall2021nerf}.
Consider a scalar input $x\in\mathbb{R}$. The first encoding is expressed as
\[
\gamma_1(x) = 
\begin{bmatrix}
\sqrt{2}\cos(f_1\,x)\,
\dots\,
\sqrt{2}\cos(f_{L}\,x)
\end{bmatrix}^{T},
\]
where the frequencies $f_i,\,i=1,\dots,L$ are drawn from a zero-mean Gaussian distribution with standard deviations specified by a custom \emph{bandwidth} hyperparameter (results for the test cases in Section~\ref{sec:results} are computed using a bandwidth equal to 100). Note that the use of a cosine function leads to the same PE irrespective of the sign of the considered frequencies.

The second approach leverages an a-priori selected frequency range~\cite{mildenhall2021nerf} that has proven useful in recurrent and attention-based architectures~\cite{gehring2017convolutional, vaswani2017attention}. This is expressed as
\[
\gamma_2(x) = 
\begin{bmatrix}
\sin(2^{0}\,x)\, 
\dots\,
\sin(2^{L-1}\,x)\,
\cos(2^{0}\,x)\,
\dots\,
\cos(2^{L-1}\,x)
\end{bmatrix}^{T},
\]
where $L$ frequencies are selected as $\{f_1,f_2,\dots,f_L\} = \{2^0, 2^{1}, 2^{L-1}\}$. In addition, $x$ is encoded as a sequence of $L$ sine and $L$ cosine functions, resulting in twice the encoded dimensionality of the previous approach.

Another feature that might significantly affect the resulting representation, consists of a selective application of positional encoding to various groups of inputs. 
This might be appropriate for applications where the frequency spectrum is different in space and time, as discussed for an harmonic pipe flow in Section~\ref{sec:xt_pipe}.
Consider applying PE only to time in a space-time field, i.e., where $\bm{x} = [x,y,z,t]^{T}$.
For the space variables we instead apply an identity encoding, where each variable in space and time is simply repeated $L$ times, i.e., $x=[x \dots x]^{T}\in\mathbb{R}^{L}$. 
This results in
\[
\Gamma(\bm{x}): \mathbb{R}^{4}\to\mathbb{R}^{4\cdot\,L} = 
\begin{bmatrix}
x & x & \dots & y & y & \dots & z & z & \dots & \gamma_{i}(t) 
\end{bmatrix},\,\,i=\{1,2\}.
\]

\subsubsection{Trainable space-frequency positional encoding for MLP-based neural field}

Since a-priori selected frequencies may not be well adapted to the unknown spectrum of a specific field, we consider the encoding through a trainable linear layer. 
For the case where this is selectively applied to spatial variables, we have
\[
\Gamma(\bm{x}): \mathbb{R}^{4}\to\mathbb{R}^{4\cdot\,L} = 
\begin{bmatrix}
g_{1}(x) & 
g_{1}(x) & 
\dots & 
g_{2}(y) & 
g_{2}(y) & 
\dots & 
g_{3}(z) & 
g_{3}(z) & 
\dots & 
\gamma_{i}(t) 
\end{bmatrix},\,\,i=\{1,2\},
\]
where $g_{k}(x) = W_{k}^T\,x + b_{k},\,k=\{1,2,3\}$ represents a linear transformation (or a feedforward neural network with no activations and hidden layers), but can easily be generalized to arbitrary MLP architectures.
This outperforms previous identity encoding, providing more flexibility to learn \emph{adapted} representations.

\subsubsection{The SIREN architecture}\label{sec:siren}

An alternative paradigm to overcome issues with spectral bias is by adopting activation functions that are better suited to represent high-frequency features. 
In this context, commonly used activation functions such as \emph{ReLU}, \emph{SiLU}, and \emph{tanh} demonstrate strong capability in approximating complex functions. 
However, they are insufficient on their own to mitigate spectral bias and are also prone to issues such as vanishing or exploding gradients.
Periodic activation functions offer a natural solution to this problem, and are at the core of sinusoidal representation networks (SIRENs)~\cite{sitzmann2020implicit}. 
A SIREN architecture incorporates layers of the form
\[
\bm{\Phi}(\bm{x}) = \bm{W}_{n}\left(\phi_{n-1}\circ \phi_{n-2}\circ \cdots \circ \phi_{0}\right)(\bm{x}) + \bm{b}_{n}
\]
where the arbitrary $i$-th layer is expressed as 
\[
\bm{x}_{i}\to \phi_{i}(\bm{x}_{i}) = \sin\left(\bm{W}_{i}\,\bm{x}_{i} + \bm{b}_{i}\right).
\]
A SIREN representation has many advantages over conventional MLPs.  
First, SIREN is infinitely differentiable and the derivative of a SIREN representation (with respect to the network inputs) is still a SIREN representation. 
This is shown in~\cite{sitzmann2020implicit} where images are reconstructed from their gradients.
This has also been demonstrated through the ability of SIREN representations to satisfy partial differential equations through PINN-based (see~\cite{karniadakis2021physics}) augmentation of the loss function.
Additionally, it can be shown that the propagation of information through a SIREN network is stable under the initialization $w_{i}\sim\mathcal{U}(-\sqrt{6/n},\sqrt{6/n})$ where $n$ is the input dimensionality.
SIREN has been demonstrated through applications in image, video, and audio representations, and in solving complex boundary value problems, an example being the Eikonal equation leading to an enhanced representation of 3D geometries~\cite{sitzmann2020implicit}. This closely aligns with yielding signed distance functions which are crucial for implicit neural networks to learn three-dimensional shape representations for geometry reconstruction.
Finally, in our implementation, the first layer uses an activation function of the form $\sin\left(\omega_{0}\cdot\bm{W}_{i}\,\bm{x}_{i} + \bm{b}_{i}\right)$ spanning multiple periods over the interval $[-1, 1]$. We use $\omega_{0}=30$ in all our tests~\cite{sitzmann2020implicit}.

\subsubsection{Multiplicative Filter Networks}\label{sec:mfn}

An alternative paradigm is offered by the concept of \emph{multiplicative filter network} (MFN~\cite{fathony2021multiplicative}) that provides a compelling performance with a much simpler architecture. 
MFN consists of a linear combination of nonlinear kernel functions $\bm{g}(\cdot)$ expressed as
\begin{equation}\label{equ:mfn}
\begin{split}
\bm{z}^{(1)} = \bm{g}\left(\bm{x};\bm{\theta}^{(1)}\right),\,\,
\bm{z}^{(i+1)} &= \left(\bm{W}^{(i)}\,\bm{z}^{(i)} + \bm{b}^{(i)}\right)\odot \bm{g}\left(\bm{x};\bm{\theta}^{(i+1)}\right),\,\,i=1,\dots,k-1,\\
\bm{\Phi}(\bm{x}) &= \bm{W}^{(k)}\,\bm{z}^{(k)} + \bm{b}^{(k)},
\end{split}
\end{equation}
where $\odot$ indicates the Hadamard product, and the choice of $\bm{g}(\cdot)$ can significantly affect the properties of the resulting representation. 
In this study, we focus on two types of MFNs, \emph{MFN-Fourier} and \emph{MFN-Gabor}. It can be shown that the representation obtained from these two architectures is equivalent to that generated by a linear combination of sinusoidal (Fourier) or Gabor wavelet bases. 
MFN-Fourier networks are obtained using a sinusoidal filter of the form 
\[
\bm{g}\left(\bm{x};\bm{\theta}^{(i)}\right) = \sin(\bm{\omega}^{(i)}\,x + \bm{\phi}^{(i)}).
\]
A common limitation shared by time-invariant spectral representations is their struggle to capture local features. 
Unlike MFN-Fourier networks, MFN-Gabor networks use Gabor filters to effectively capture both frequency and spatial locality. 
The Gabor kernel for the $j$-th feature in the $i$-th layer is defined as
\[
g_{j}\left(\bm{x};\bm{\theta}^{(i)}\right) = \exp\left(-\frac{\bm{\gamma}^{(i)}_{j}}{2}\Vert \bm{x}- \bm{\mu}^{(i)}_{j}\Vert^{2}_{2}\right)\,\sin(\bm{\omega}_{j}^{(i)}\,\bm{x} + \bm{\phi}_{j}^{(i)}).
\]
Additionally, it is shown that the representation in~\eqref{equ:mfn} is equivalent to a Fourier or Gabor series expansion with a number of terms that \emph{increases exponentially} with the depth of the network~\cite{fathony2021multiplicative}.

\subsubsection{Instant NGP with multi-resolution hash encoding}\label{sec:mhe}

\begin{figure}[!ht]
\centering
\includegraphics[width=\textwidth]{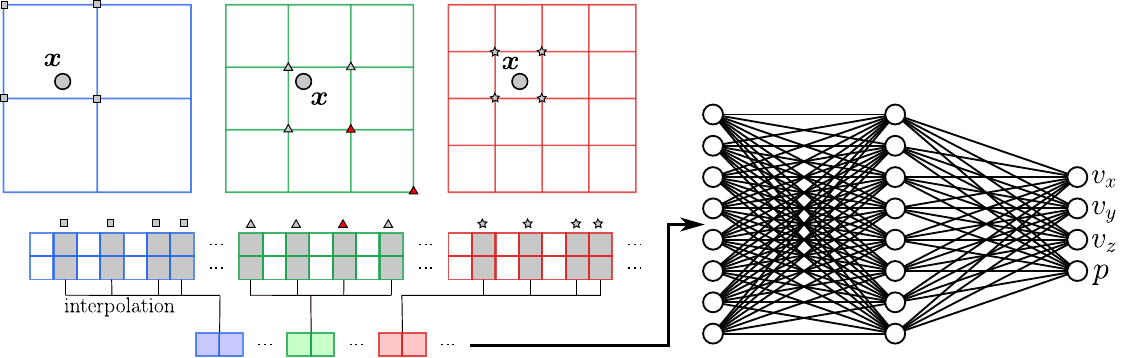}
\caption{Schematic of multiresolution hash encoding (MHE).}
\label{fig:mhe_scheme}
\end{figure}

Multi-resolution hash encoding (MHE~\cite{muller2022instant}) provides a trainable encoding defined on multiple hierarchically-arranged grids that is adapted to the target field, and allows smaller neural networks to be used without sacrificing accuracy. 
The hash table's multiresolution design helps disambiguate hash collisions, which enhances adaptability and efficiency.
A sketch with the sequence of operations performed by MHE is illustrated in Figure~\ref{fig:mhe_scheme}.

Encoding is accomplished by finding the cells that contain a given location $\bm{x}$ in a hierarchical representation with $L$ levels,
selected in a range of resolutions $[R_{min}, R_{max}]$ from the coarsest to the finest, with an arbitrary number of subdivisions from one level to the next.
Each level in the hierarchy is associated with a hash table $H_{i} \in \mathbb{R}^{T \times D},\,i = 1, 2, \dots, L$, representing $T$ feature vectors with dimensionality $D$. The hash table stores the feature vectors that refer to the vertices of the grid partition containing $\bm{x}$. These feature vectors are then interpolated based on the relative position of $\bm{x}$ with respect to such vertices. 
Interpolated coordinates are then concatenated producing a final input $\bm{\phi} \in \mathbb{R}^{L\cdot D + a}$, where $a$ represents a number of auxiliary input variables.
The study in~\cite{muller2022instant} suggests the use of hyperparameters $[L,D]=[16,2]$ as a default setting, as it produces optimal results for a variety of applications. Additional hyperparameters such as $R_{max}$ and $log(T)$ can also be fine-tuned for improved performance, see, e.g., Section~\ref{sec:xyzt_aorta}.

Finally, MHE is designed to allow rapid training of neural graphics primitives by using smaller neural networks. In this context, significant speedups in training have been demonstrated using  efficient implementations with fully-fused CUDA kernels and minimizing wasted bandwidth and compute operations.
This improvements lead to training times of seconds for images with resolution 1920x1080~\cite{muller2021real}. 
In this paper, we use the implementation provided by the \emph{PhysicsNemo} and \emph{tiny-cuda-nn} libraries from NVIDIA, available at \url{https://github.com/NVIDIA/physicsnemo}, and \url{https://github.com/NVlabs/tiny-cuda-nn}, respectively.

\subsubsection{Hyperparameter selection}\label{sec:hyper_choice}

We used the following hyperparameters for all networks. The baseline number of iterations is equal to 10,000, with a batch size of 1024 and a constant learning rate equal to $1.0\times10^{-4}$. All network architectures include 5 layers, each with 512 neurons. We use \emph{Tanh} activations for MLP-based architectures and \emph{SiLU}~\cite{elfwing2018sigmoid} for MHE. Additionally, Baseline MHE settings included a number of grid levels $L$ equal to 16, a logarithmic size of the hash map $log(T)$ of 19, a dimensionality of the feature vector $D$ equal to 2, and a maximum resolution $R_{max}$ of 32. 
Additional hyperparameter choices are reported in the table in Section~\ref{sec:hyper}.

\subsubsection{Interpolation and memorization}\label{sec:interp_memoriz}

It should be noted that all methods discussed above, including positional encoding, sinusoidal activations, multifilter networks and MHE are designed to \emph{promote overfit} in the early phases of training, thus increasing the flexibility of the neural representation. 
This is equivalent to boost the ability of a network to \emph{memorize} the values of a field at a number of pre-defined training locations.
At the same time, overfitting leads to poor interpolation, i.e., poor generalization outside the input locations seen at training.
Note that this makes the above approaches not ideal in scenarios characterized by training datasets of limited size (or the so-called \emph{small-data} regime). 
We discuss this aspect further in Section~\ref{sec:overfit_vs_interp}.
To overcome this issue and retain both the memorization and interpolation properties of the representation, we train neural representations of pressure and velocity fields \emph{at the nodes of an unstructured mesh} of linear (4 node) tetrahedral elements. 
We then rely on shape function interpolation for local generalization, result extraction and visualization.

\subsection{Signed Distance Fields}\label{sec:sdf_dataset}

\begin{figure}[!ht]
\centering
\includegraphics[width=0.6\textwidth]{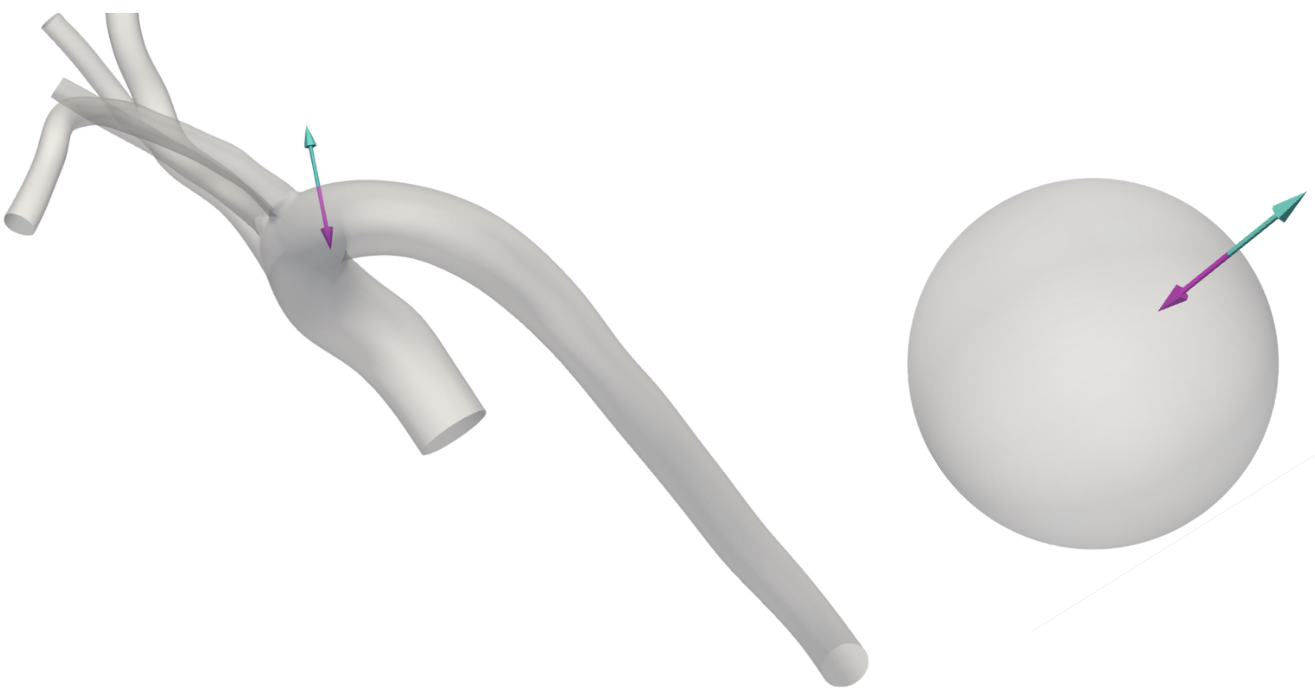}
\caption{Spatial locations outside of geometries are positive (blue), whereas locations inside are negative (pink). Points on the surface form the zero level set.}
\label{fig:sdf_normal}
\end{figure}

Implicit neural representations can also be used to represent three-dimensional geometries, through signed distance fields (SDF~\cite{park2019deepsdf}). 
Essentially, an orientable surface can be represented as the \emph{zero levelset} of a scalar distance field. 
Thus, all spatial locations belonging to the surface are associated to a zero scalar, whereas locations the are internal and external with respect to the surface are characterized by a negative and positive scalar, respectively. This is consistent with the usual convention, in computational mechanics, of having a positive normal pointing outwards, as shown in Figure~\ref{fig:sdf_normal}.
Alternatively, occupancy networks~\cite{mescheder2019occupancy} represent the three-dimensional surfaces as the decision boundary of a neural network classifier, offering an alternative to SDFs. 
While occupancy networks are often simpler to train, SDF-based approaches learn the distance to the surface, which can be useful for tasks like collision detection. Furthermore, learning a continuous field allows for smooth interpolation, seamless computation of the normal vector to the field, and simpler computation of a surface mesh by extracting the contour at zero.

Additionally, neural fields offer a number of advantages over traditional mesh-based representations. 
Reconstructions are continuous, they achieve substantial compression rates, and can be trained from experiments, simulations, or constrained to satisfy partial differential equations~\cite{sitzmann2020implicit}. 
They can be formulated to characterize reconstruction uncertainty~\cite{srivastava2014dropout,ran2023neurar}, and to encode interfaces and complex geometries. 
However, they carry some limitations with certain topological details, for example, thin or extremely fine structures (e.g. hair-like geometries).

\subsection{Datasets}

We first discuss in detail the datasets generated for the experiments presented in the next section. 
A summary of all datasets used in this paper is reported in Table~\ref{tab:all_ds}.
\begin{table}[!ht]
\centering
\caption{Dataset Summary}
\resizebox{\textwidth}{!}{
\begin{tabular}{l l c c}
\toprule
{\bf Neural Field Type} & {\bf Space-Time} & {\bf Datasets Type} & {\bf Description} \\
\midrule
{\bf Pressure/velocity} & XT & Cylindrical pipe & 1D space-time results from simulation\\
{\bf Image}  & XY & Greyscale image & 2D image reconstruction/compression\\
{\bf Pressure/velocity} & XYZ & Cylindrical pipe & Steady simulation results\\
{\bf Pressure/velocity} & XYZ & Patient-specific aorta & Steady simulation results\\
{\bf Pressure/velocity} & XYZT & Cylindrical pipe & Pulsatile simulation results\\
 {\bf Pressure/velocity} & XYZT & Patient-specific aorta & Pulsatile simulation results\\
 \midrule
{\bf Signed distance} & SDF(XYZ) & Sphere, pulmonary tree, thoracic aorta & Ideal shapes and patient-specific anatomies\\
{\bf Signed distance} & SDF(XYZ) & Thoracic aortic zoo & Library of patient-specific anatomies\\
\bottomrule
\end{tabular}}
\label{tab:all_ds}
\end{table}

\subsubsection{Pressure/Velocity Field Reconstruction}\label{sec:ds_field}

We list the datasets used in the numerical experiments for the reconstruction of two-dimensional images and pressure/velocity fields from cardiovascular simulation.

\vspace{5pt}

\begin{figure}[!ht]
\centering
    \begin{subfigure}[b]{0.28\textwidth}
        \includegraphics[width=\textwidth]{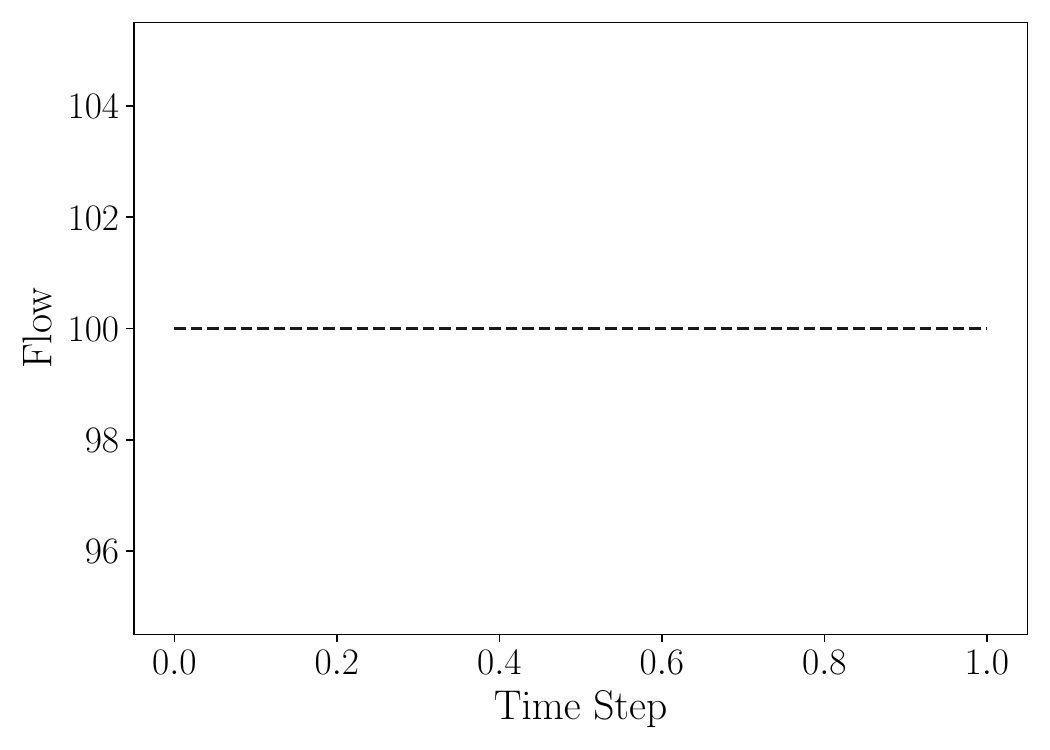}
        \caption{Steady flow profile}
        \label{fig:steady_flow_pipe}
    \end{subfigure}
    \begin{subfigure}[b]{0.28\textwidth}
        \centering
        \includegraphics[width=\textwidth]{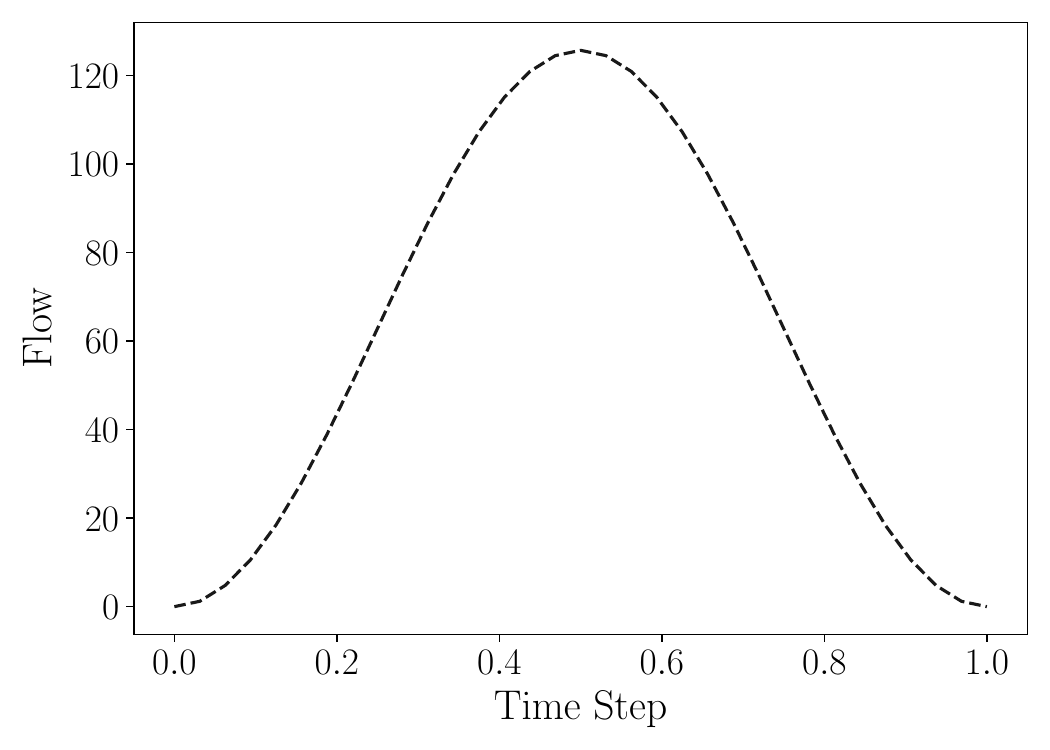}
        \caption{Parabolic flow profile}
        \label{fig:harmonic_flow_pipe}
    \end{subfigure}
    \begin{subfigure}[b]{0.28\textwidth}
        \centering
        \includegraphics[width=\textwidth]{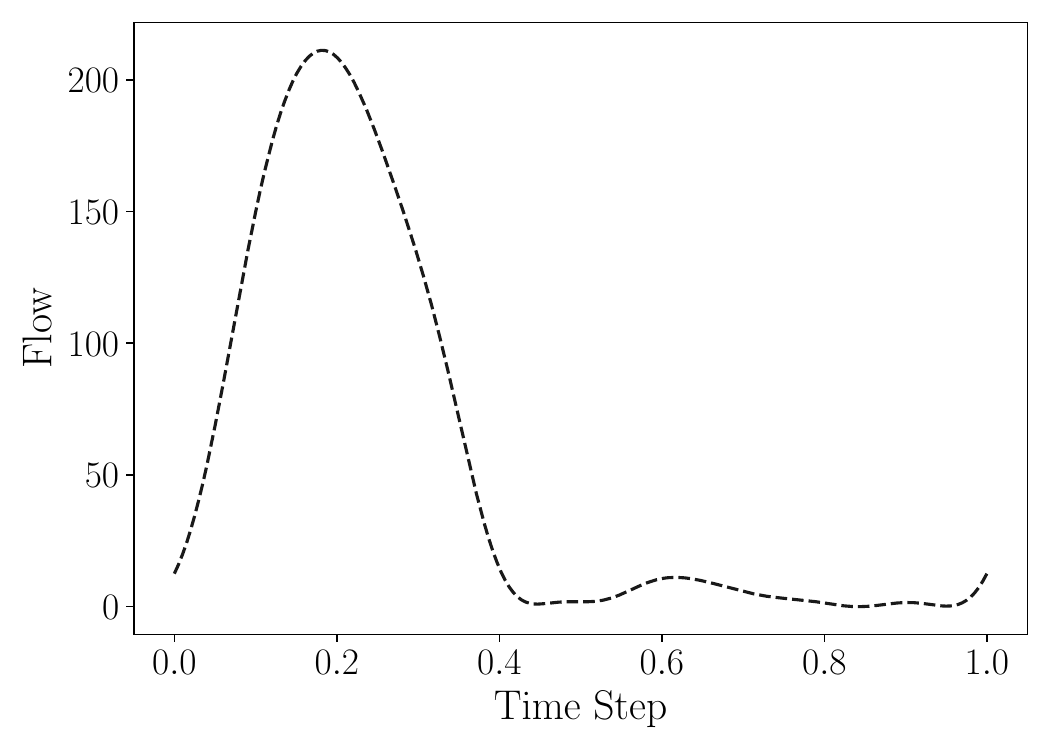}
        \caption{Pulsatile aortic flow profile}
        \label{fig:pulsatile_aorta}
    \end{subfigure}
\caption{Flow profiles in mL/s for pipe and aorta test cases.}
\label{fig:pipe_aorta_flow}
\end{figure}

\noindent{\bf Pipe} - The cylindrical pipe geometry is an ideal aorta centered at the origin with a radius of 2 cm and the length being 30 cm along the Z-axis. This three dimensional geometry consists of a polygonal surface mesh composed of triangular elements. The mesh is bounded within the domain $(x, y, z) \in [-2, 2] \times [-2, 2] \times [0, 30]$. The mesh contains 11,208 tetrahedral elements, 2,354 nodes and 160 time steps.
We perform two types of simulation, a steady state simulation with constant flow shown in Figure~\ref{fig:steady_flow_pipe}, and a simulation with the harmonic flow in Figure~\ref{fig:harmonic_flow_pipe}, both with a parabolic (Poiseuille-like) spatial profile. 
Simulation parameters and boundary conditions are also reported in Table~\ref{tab:pipe_simul_params}.
We run the simulations using the \texttt{svMultiPhysics} solver from the SimVascular software platform (\url{https://github.com/SimVascular/svMultiPhysics}).

\vspace{5pt}

\noindent{\bf Greyscale Image} - The grayscale image of a zebra has dimensions of $256 \times 186$ pixels with a file size of 58,487 bytes. It was selected for the significant variations in contrasts due to the skin strides and in the background. 

\vspace{5pt}

\noindent{\bf Aorta} - The patient-specific aortic anatomy considered in this study was downloaded as model \texttt{0011\_H\_AO\_H}, from the Vascular Model Repository (VMR, \url{https://www.vascularmodel.com/dataset.html}). 
This anatomy was acquired from a MR scan of a healthy 23 year old female and used as a control in comparative healthy-diseased studies~\cite{jr2011computational}.
The triangular surface mesh consists of 158,652 cells and 79,328 points, and has bounds $(x, y, z) \in [-8.276,4.23732] \times [-2.49448, 4.20527] \times [-15.3465, 12.216]$. 
The model was solved under open loop boundary conditions and pulsatile aortic inflow as shown in Figure~\ref{fig:pipe_aorta_flow}. RCR boundary conditions were also applied and five outlets, as reported in Table~\ref{tab:aorta_rcr}.

\begin{figure}[!ht]
\centering
    \begin{subfigure}[b]{0.4\textwidth}
        \centering
        \includegraphics[width=0.8\textwidth]{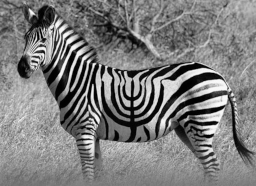}
        \caption{Greyscale Image}
    \end{subfigure}
    \begin{subfigure}[b]{0.07\textwidth}
        \includegraphics[width=\textwidth]{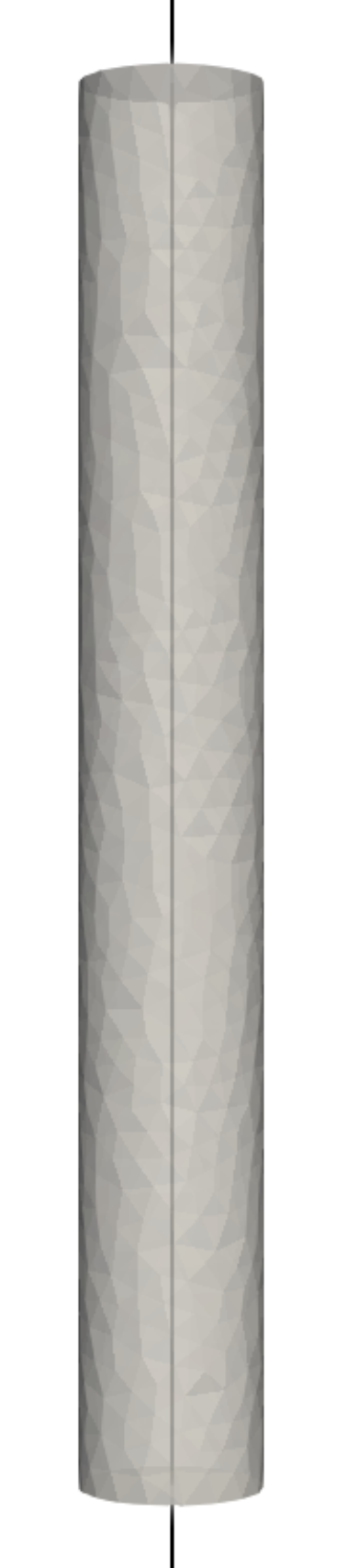}
        \caption{Pipe}
    \end{subfigure}
    \begin{subfigure}[b]{0.12\textwidth}
        \centering
        \includegraphics[width=\textwidth]{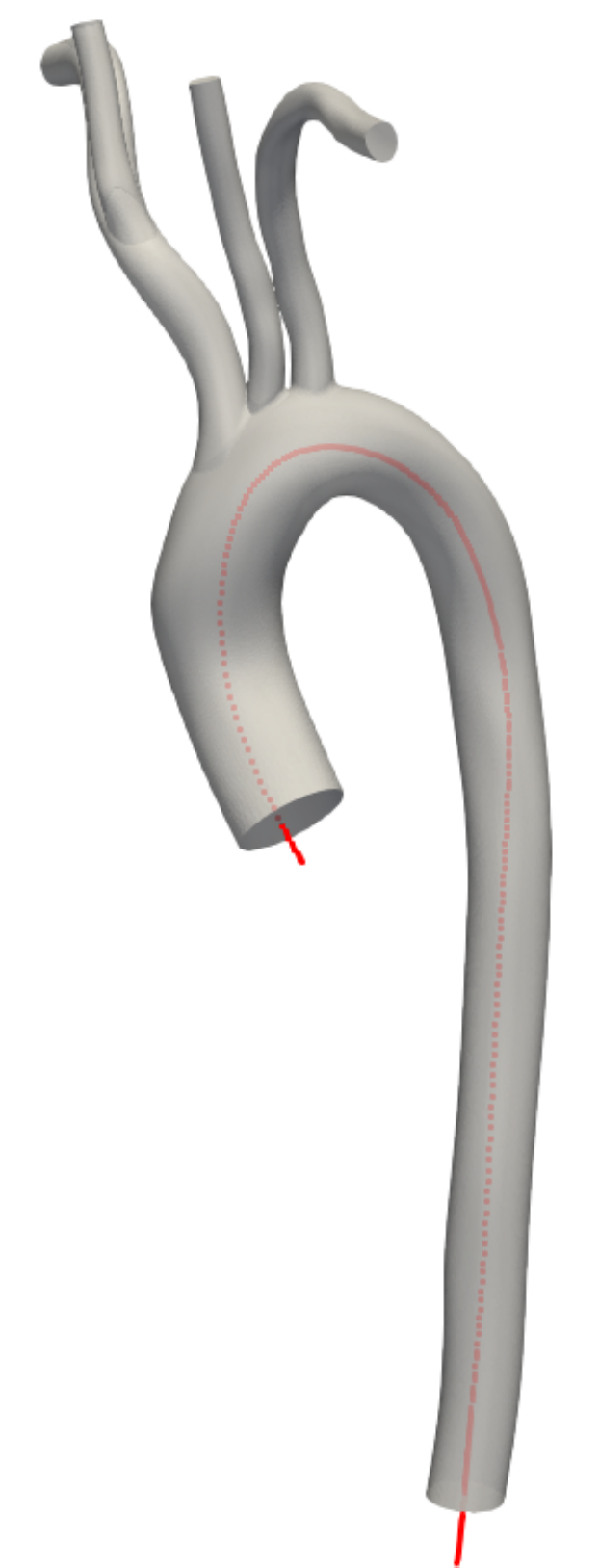}
        \caption{Aorta}
    \end{subfigure}
\caption{Visualization of examples from the selected test cases: greyscale zebra image, cylindrical pipe, and aortic anatomy.}
\end{figure}

\begin{table}[!ht]
\centering
\caption{Simulation Parameters - XYZ Pipe.}
\label{tab:pipe_simul_params}
\resizebox{\textwidth}{!}{%
\begin{tabular}{c c c c c c}
\toprule
{\bf Time Step} (s) & {\bf Number of Time Steps} & $\boldsymbol{R_{p}}$ (dynes/s/cm$^{-5}$) & $\boldsymbol{C}$ (cm$^5$/dynes) & $\boldsymbol{R_{d}}$ (dynes/s/cm$^{-5}$) & $\boldsymbol{P_{d}}$ (mmHg) \\
\midrule
 0.005 & 1600 & 121.0 & $1.5\times 10^{-4}$ & 1212.0 & 0.0 \\
 \bottomrule
\end{tabular}}
\end{table}

\begin{table}[!ht]
\centering
\caption{RCR Boundary Condition Parameters - XYZ Aorta.}
\begin{tabular}{l c c c c}
\toprule
{\bf Branch} & $\boldsymbol{R_{p}}$ (dynes/s/cm$^{-5}$) & $\boldsymbol{C}$ (cm$^5$/dynes) & $\boldsymbol{R_{d}}$ (dynes/s/cm$^{-5}$) & $\boldsymbol{P_{d}}$ (mmHg)\\
\midrule
{\bf Brachiocephalic Trunk} & 1011.0 & 9.149e-05 & 17038.0 & 0.0\\
{\bf Right Carotid Artery} & 1523.0 & 6.072e-05 & 25670.0 & 0.0\\
{\bf Carotid} & 1523.0 & 6.072e-05 & 25670.0 & 0.0\\
{\bf Subclavian} & 1237.0 & 7.477e-05 & 20847.0 & 0.0\\
{\bf Aortic outlet} & 200.0 & 0.00046231 & 3372.0 & 0.0\\
\bottomrule
\end{tabular}
\label{tab:aorta_rcr}
\end{table}

\subsubsection{Signed Distance Fields}\label{sec:ds_sdf}

We now list the datasets used in the numerical experiments for the reconstruction of idealized and patient-specific anatomies with signed distance fields. 

\vspace{5pt}

\noindent{\bf Sphere} - We consider an ideal spherical geometry consisting of a sphere centered at the origin with a radius of 0.5 cm, discretized through a triangular surface mesh containing 19,600 cells and 9,802 points, and having bounds $(x, y, z) \in [-0.5, 0.5] \times [-0.5, 0.5] \times [-0.5, 0.5]$ cm.

\vspace{5pt}

\noindent{\bf Pulmonary anatomy} - A pulmonary anatomy was downloaded from the Vascular Model Repository (VMR, \url{https://www.vascularmodel.com/dataset.html}) from a 0.2 year old female patient, diagnosed with Williams Syndrome, characterized by peripheral pulmonary artery stenosis (PPAS). 
This model was previously used in a study of virtual transcatheter intervention, with further details on the surgical procedure and post processing available from~\cite{lan2022virtual}. 
The triangular surface mesh of the pre-stent anatomy consist of 351,518 cells and 175,761 points and has bounds $(x, y, z) \in [-5.27724,2.30128] \times [-11.4598, -5.53834] \times [-12.0477,-6.06122]$ cm.

\vspace{5pt}

\noindent{\bf Aortic anatomy} - We consider the same aortic anatomy we have introduced in the previous section.

\vspace{5pt}

\begin{figure}[!ht]
    \centering
    \begin{minipage}[b]{0.49\textwidth}
        \centering
        \begin{subfigure}[b]{\textwidth}
            \centering
            \includegraphics[width=0.5\textwidth]{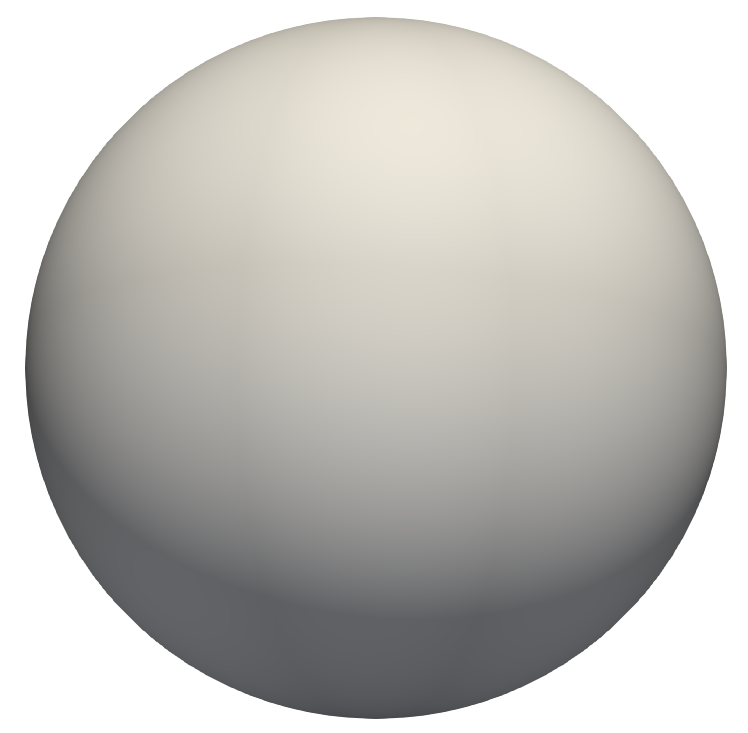}
            \caption{Sphere}
            \label{fig:sphere}
        \end{subfigure}
        \vspace{0.1cm}
        \begin{subfigure}[b]{\textwidth}
            \centering
            \includegraphics[width=0.7\textwidth]{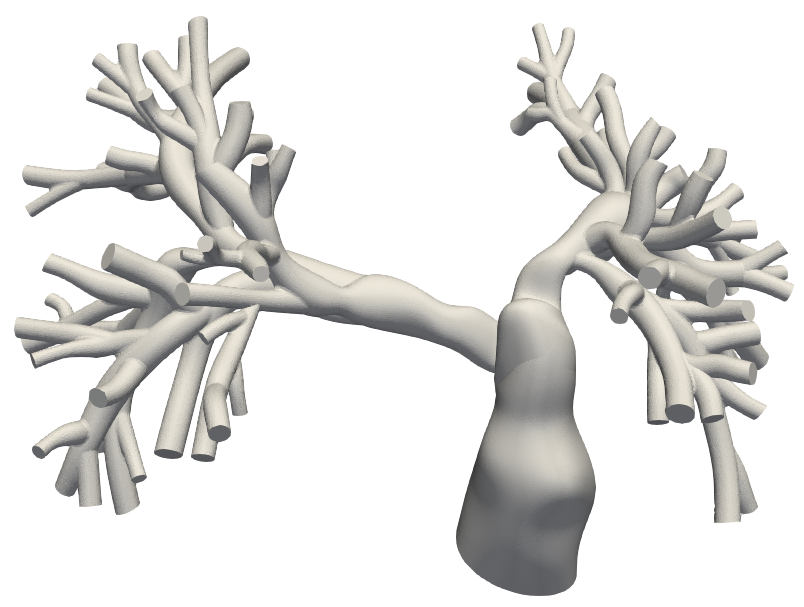}
            \caption{Pulmonary}
            \label{fig:pulm}
        \end{subfigure}
    \end{minipage}
    \begin{minipage}[b]{0.49\textwidth}
        \centering
        \begin{subfigure}[b]{\textwidth}
            \centering
            \includegraphics[width=0.32\textwidth]{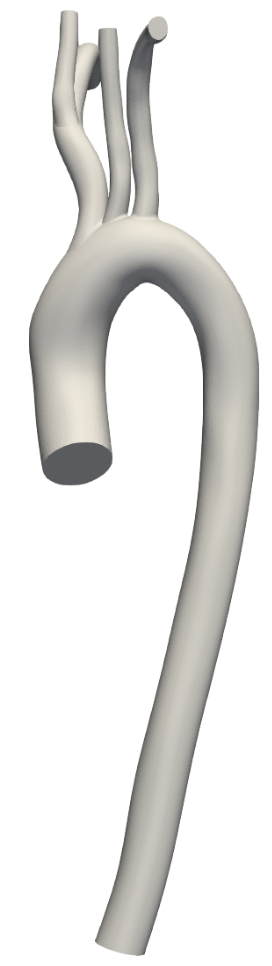}
            \caption{Aorta}
        \end{subfigure}
    \end{minipage}
    \caption{Selected test cases: ideal spherical geometry, pulmonary, and aortic anatomies.}
    \label{fig:datasets}
\end{figure}

\begin{figure}[!ht]
\centering
    \begin{subfigure}[b]{0.33\textwidth}
    	\centering
    	\includegraphics[width=\textwidth]{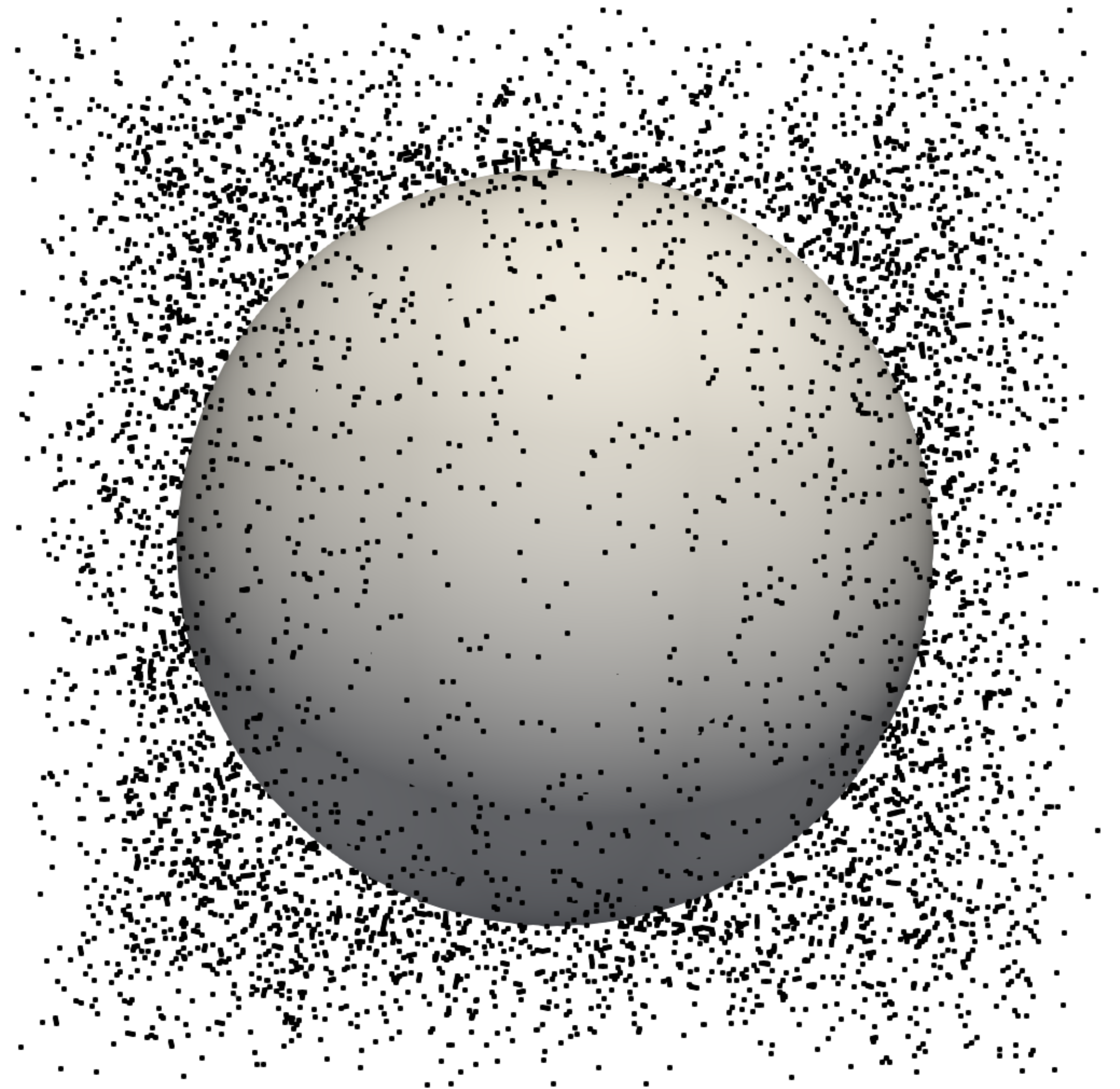}
    	\caption{Uniform}
    \end{subfigure}
    \begin{subfigure}[b]{0.31\textwidth}
        \centering
        \includegraphics[width=\textwidth]{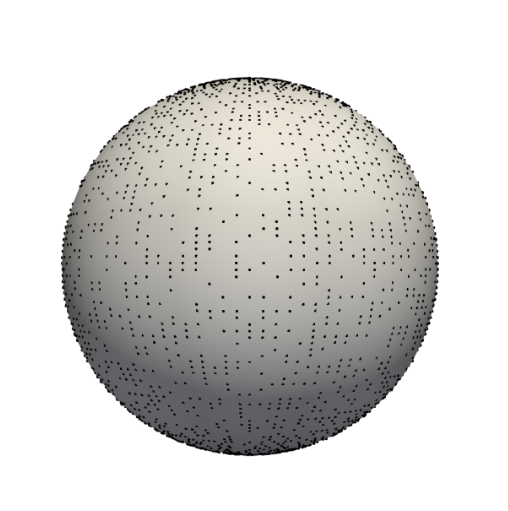}
        \caption{Surface}
    \end{subfigure}
    \begin{subfigure}[b]{0.3\textwidth}
        \centering
        \includegraphics[width=\textwidth]{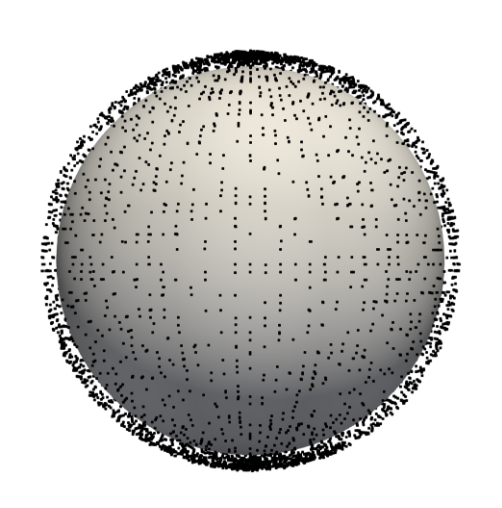}
        \caption{Perturbed}
    \end{subfigure}
\caption{The three classes of training samples for SDF-based geometry learning.}
\label{fig:sdf_sample_class}
\end{figure}

We first rescale the model geometry so it fits within the three-dimensional unit cube $[0,1]^{3}$. 
We then partition the SDF dataset consisting of pairs $(\bm{x}_{i},d_{i})_{i=1}^{N}$ in three classes with number of samples equal to $N_1$, $N_2$ and $N_3$, respectively, such that $N_1 + N_2 + N_3 = N$.
The first class includes $N_1$ \emph{uniform} samples drawn from $[0,1]^{3}$. For each point, the signed distance function is computed using the VTK library by first converting the triangulated surface representing the anatomy in a \texttt{polydata} data structure, and then applying the \texttt{vtkImplicitPolyDataDistance} filter. 
The second class includes $N_2$ \emph{surface} points for which the SDF is known and equal to zero. These points are drawn on the surface triangulation proportionally to the area of each triangle. 
Finally, we consider $N_3$ \emph{perturbed} points, randomly sampled on each surface triangle, but then also randomly perturbed in the normal direction following a zero-mean Gaussian distribution with standard deviation $\sigma = r/\delta_{\text{SDF}}$. 
The radius of the rescaled model is assumed as $r=1/2$, while numerical tests are conducted by setting the factor $\delta_{\text{SDF}}=1024$. We examined the effects of varying the perturbation factor $\delta_{\text{SDF}}$ from the baseline value $\alpha = 1024$, i.e., for \emph{perturbed} training datasets that are closer ($\alpha = 5096$) or farther away ($\alpha = 256$) from the surface mesh. In terms of mean distance error, perturbation levels do not appear to have a dominant influence in determining the quality of geometry representation. 
A visual representation of the samples for each of these three classes is shown in Figure~\ref{fig:sdf_sample_class}.

Moreover, we test with six variations (see Table~\ref{tab:sdf_large_small}) to understand which type of sample (uniform, surface or perturbed) mostly affects reconstruction accuracy. 
First, we test with a large number of samples (total of 580K) partitioned as: 500K uniform samples, 40K surface samples, and 40K perturbed samples (roughly 86\%, 7\%, and 7\%). We then rotate these three types so that each type has 500K samples and the other two have 40K.
We also consider a second scenario (which we denote as \emph{small}, while referring to previous scenario as \emph{large}) where the number of samples is equal to 100K, 8K, and 8K applied to all sample types, see Table~\ref{tab:sdf_large_small}.

\begin{table}[!ht]
\centering
\caption{Scenarios for SDF training dataset generation.}
\resizebox{.5\textwidth}{!}{%
\begin{tabular}{l c c c c}
\toprule
& & \multicolumn{3}{c}{\bf Sample types}\\
\multicolumn{2}{c}{\bf Scenario} & {\bf Uniform} & {\bf Surface} & {\bf Perturbed}\\
\midrule
\multirow{3}{*}{\bf LARGE} & {\bf MSS} & 500K & 40K & 40K  \\
 & {\bf SMS} & 40K & 500K & 40K  \\
 & {\bf SSM} & 40K & 40K & 500K  \\
\midrule
\multirow{3}{*}{\bf SMALL} & {\bf MSS} & 100K & 8K & 8K \\
 & {\bf SMS} & 8K & 100K & 8K \\
 & {\bf SSM} & 8K & 8K & 100K \\
\bottomrule
\end{tabular}}
\label{tab:sdf_large_small}
\end{table}

\subsubsection{Thoracic Aorta Zoo}

\begin{figure}[!ht]
\centering
\begin{subfigure}[b]{0.9\textwidth}
	\includegraphics[width=\textwidth]{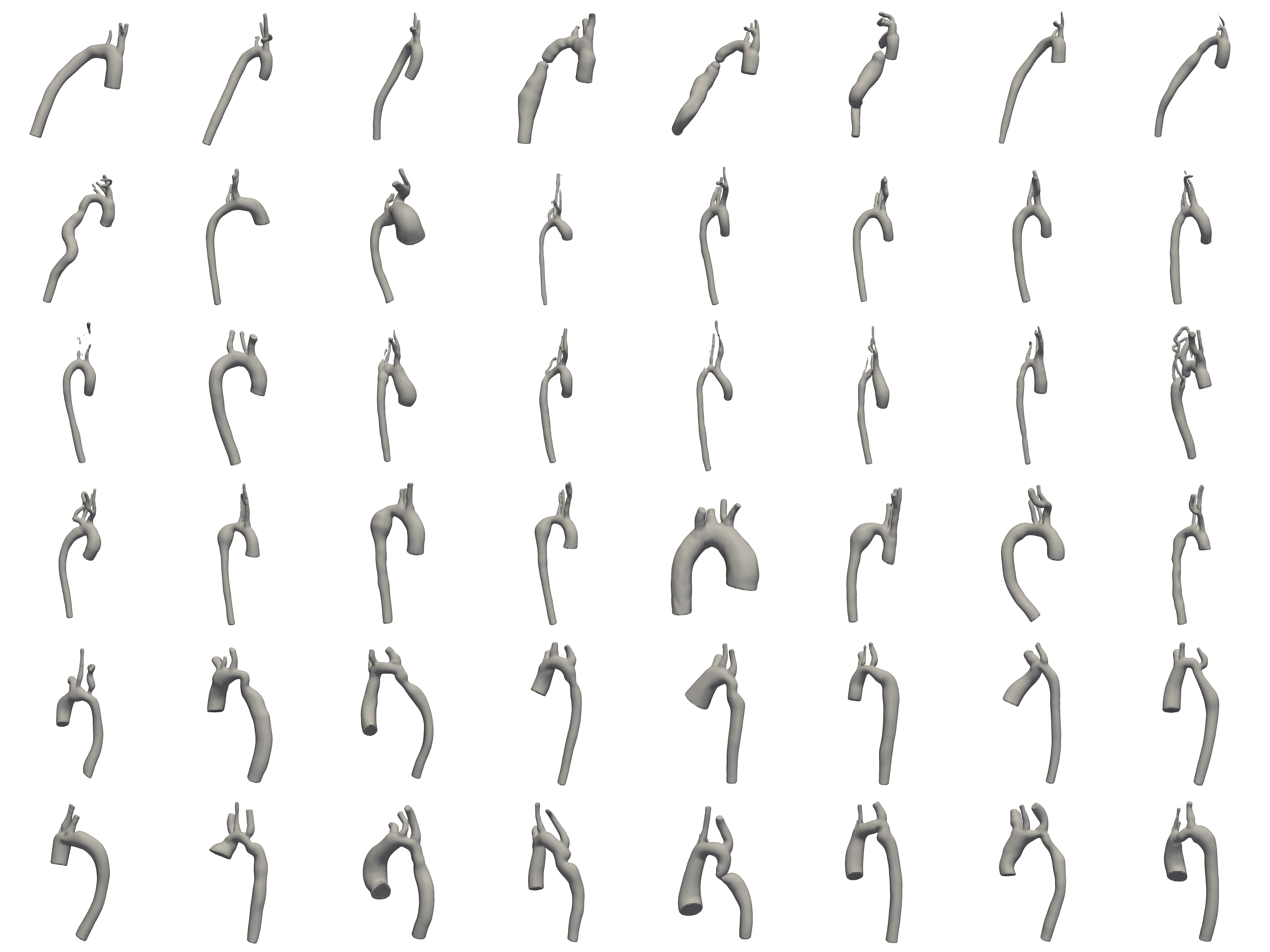}
\end{subfigure}
\caption{\emph{Zoo} of thoracic aortic anatomies from the vascular model repository (at \url{https://www.vascularmodel.com/}). A list is provided in Table~\ref{tab:models_aortic_zoo}.}
\label{fig:sdf_zoo_all_models}
\end{figure}

We further extend our SDF investigation considering a \emph{zoo} of 48 thoracic aortic anatomies from the Vascular Model Repository (the full list of models is reported in Table~\ref{tab:models_aortic_zoo}). A visualization of the entire model zoo is provided in Figure~\ref{fig:sdf_zoo_all_models}.

\begin{table}[!ht]
\centering
\caption{List of model from VMR in aortic zoo.}
\label{tab:zoo_model_list}
\resizebox{\textwidth}{!}{
\begin{tabular}{c | c c c c c c}
\toprule
 & {\bf 1} & {\bf 2} & {\bf 3} & {\bf 4} & {\bf 5} & {\bf 6}\\
\midrule
{\bf 1} & 0001\_H\_AO\_SVD & 0002\_H\_AO\_SVD & 0003\_H\_AO\_SVD & 0004\_H\_AO\_SVD & 0005\_H\_AO\_SVD & 0006\_H\_AO\_SVD \\
{\bf 2} & 0007\_H\_AO\_H   & 0008\_H\_AO\_H   & 0009\_H\_AO\_H   & 0010\_H\_AO\_H   & 0011\_H\_AO\_H   & 0012\_H\_AO\_H \\
{\bf 3} & 0013\_H\_AO\_COA & 0014\_H\_AO\_COA & 0015\_H\_AO\_COA & 0016\_H\_AO\_COA & 0017\_H\_AO\_COA & 0018\_H\_AO\_COA \\
{\bf 4} & 0019\_H\_AO\_COA & 0020\_H\_AO\_COA & 0021\_H\_AO\_MFS & 0022\_H\_AO\_MFS & 0023\_H\_AO\_MFS & 0024\_H\_AO\_H \\
{\bf 5} & 0025\_H\_AO\_MFS & 0026\_H\_AO\_MFS & 0027\_H\_AO\_MFS & 0093\_A\_AO\_H   & 0094\_A\_AO\_H   & 0095\_A\_AO\_H \\
{\bf 6} & 0096\_A\_AO\_COA & 0097\_A\_AO\_COA & 0098\_A\_AO\_COA & 0099\_A\_AO\_COA & 0100\_A\_AO\_COA & 0101\_A\_AO\_COA \\
{\bf 7} & 0221\_H\_AO\_AOD & 0222\_H\_AO\_AOD & 0223\_H\_AO\_AOD & 0224\_H\_AO\_AOD & 0225\_H\_AO\_COA & 0226\_H\_AO\_COA \\
{\bf 8} & 0227\_H\_AO\_COA & 0228\_H\_AO\_COA & 0229\_H\_AO\_COA & 0230\_H\_AO\_COA & 0231\_H\_AO\_COA & 0232\_H\_AO\_COA \\
\bottomrule
\end{tabular}}
\label{tab:models_aortic_zoo}
\end{table}

\section{Results}\label{sec:results}

\subsection{List of INR Networks}

As introduced in Section~\ref{sec:inr_architectures}, we investigate the performance of various architectures.
First we consider a \emph{vanilla} MLP architecture, without PE, special activation functions, kernel or hash encoding, as a baseline for comparison.
Next we implement PE with frequencies randomly drawn from a zero-mean Gaussian distribution with a default bandwidth of 100, and passed through a cosine function. This configuration is denoted as \emph{MLP PE}. 
We also consider the \emph{MLP PE 2L} variant, where $L$ frequencies are fixed, non-trainable, encoded using both sine and cosine, resulting in an encoding of size $2L$.
Furthermore, in scenarios where it may be advantageous to retain linear representations for specific input components, we enable selective encoding strategies. 
These include an identity mapping for the spatial coordinates (\emph{MLP PE 2L ID}) and an architecture where the spatial coordinates are encoded through a trainable linear layer (\emph{MLP PE 2L LIN}). 
Finally, we consider SIREN, multifilter networks and MHE with the architectures discussed in Section~\ref{sec:siren}, \ref{sec:mfn}, and \ref{sec:mhe}, respectively. 
Hyperparameters for all architectures are discussed in Section~\ref{sec:hyper_choice} and~\ref{sec:hyper}.
When learning fields with multiple components, the loss function is evaluated using independently normalized output components, and the outputs are rescaled back to their original ranges at validation.
The INR architectures adopted in this study are listed in Table~\ref{tab:all_inr_architectures}.
\begin{table}[!ht]
\centering
\caption{Network architectures compared in this study.}
\resizebox{.7\textwidth}{!}{%
\begin{tabular}{l c}
\toprule
{\bf Model Name} & {\bf Description}\\
\midrule
{\bf MLP} & Multilayer Perceptron \\
\midrule
{\bf MLP PE} & MLP with Positional Encoding (Random Frequency) \\
{\bf MLP PE 2L} & MLP with Positional Encoding (Fixed Frequency) \\
{\bf MLP PE 2L ID} & MLP PE 2L with identity space encoding \\
{\bf MLP PE 2L LIN} & MLP PE 2L with trainable linear space encoding \\
\midrule
{\bf SIREN} & Sinusoidal Representation Networks \\
\midrule
{\bf MFN-Fourier} & Multiplicative Filter Networks with Fourier Filters (FourierNets) \\
{\bf MFN-Gabor} & Multiplicative Filter Networks with Gabor Filters (GaborNets)\\
\midrule
{\bf MHE} & Multi-resolution Hash Encoding \\
\bottomrule
\end{tabular}}
\label{tab:all_inr_architectures}
\end{table}

\subsection{XT Pipe}\label{sec:xt_pipe}

\begin{figure}[!ht]
\begin{subfigure}[t]{\textwidth}
\includegraphics[width=\textwidth]{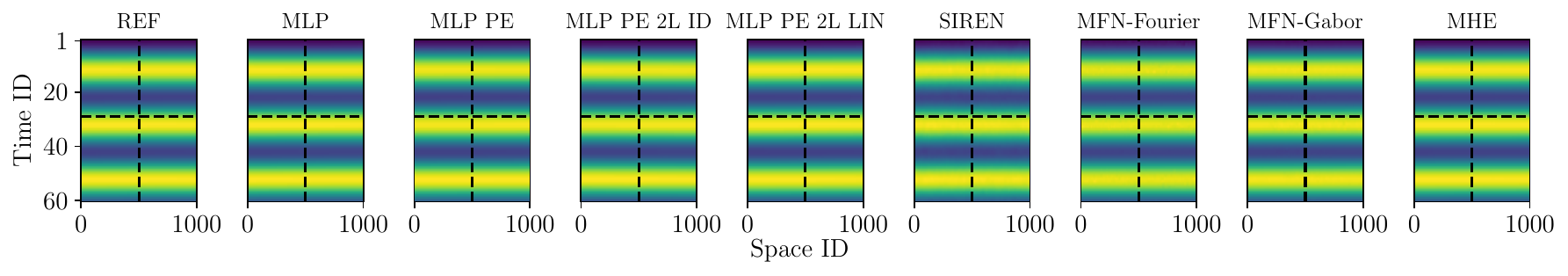}
\caption{Space-time reconstructions for the various methods.}
\label{fig:xt_pipe_2d}
\end{subfigure}

\begin{subfigure}[t]{\textwidth}
\includegraphics[width=\textwidth]{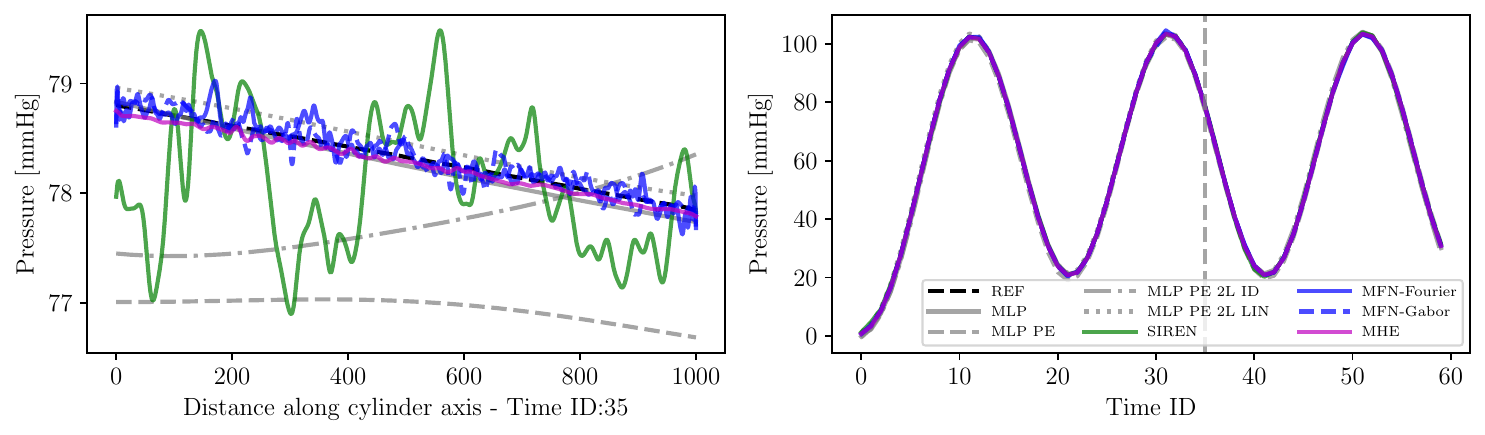}
\caption{Results at fixed time (left) and space location (right).}
\label{fig:xt_pipe_profiles}
\end{subfigure}
\caption{Results - XT Pipe.}
\label{fig:xt_pipe}
\end{figure}

In this first test case we focus on a map $f:\mathbb{R}^2 \rightarrow \mathbb{R}$ or $p=f(\bm{x}) = f(x,t)$, where $p$ represents the pressure along the cylindrical axis $x$, and $t$ is time. Values of tested hyperparameters are reported in Table~\ref{tab:xt_pipe_hyper}.
Figure~\ref{fig:xt_pipe_2d} displays all reconstructions where the spatial position along the cylinder axis is plotted against time. From visual inspection all reconstructions appear to be identical. However, a closer look reveals, on one hand, the ability of all approaches to capture the dominant oscillatory response in time, but also differences in the ability to capture a physically relevant linear pressure variation in space, typical of an oscillating Poiseuille flow. This justifies using a different encoding for spatial and temporal features.

Figure~\ref{fig:xt_pipe} shows the results obtained from the selected architectures, particularly from MLP networks with fixed, random, and trainable linear PE. 
While frequency-based spatial encoding produces spurious oscillations, linear spatial encoding mitigates this issue, allowing the networks to better adapt to the correct linear pressure profile. The RMSE errors in Table~\ref{tab:xt_pipe_rmse} support such conclusion. The best accuracy is achieved by MHE followed by MFN-Gabor and MFN-Fourier.

\begin{table}[!ht]
\centering
\caption{Pressure RMSE - XT Pipe.}
\label{tab:xt_pipe_rmse}
\begin{tabular}{l c}
\toprule
{\bf Models} & RMSE $p$ [mmHg]\\
\midrule
MLP & 6.415$\times$10$^{-1}$\\
MLP PE & 1.199\\
MLP PE 2L ID & 1.256\\
MLP PE 2L LIN & 8.088$\times$10$^{-1}$\\
SIREN & 5.181$\times$10$^{-1}$\\
MFN-Fourier & 3.560$\times$10$^{-1}$\\
MFN-Gabor & 1.983$\times$10$^{-1}$\\
{\bf MHE} & $\boldsymbol{5.483\times10^{-2}}$\\
\bottomrule
\end{tabular}
\end{table}

\subsection{XY Zebra}\label{sec:res_zebra}

We evaluate the performance of INRs when representing two-dimensional greyscale image data. 
Since most of the literature on neural fields includes examples of image reconstruction where accuracy is evaluated based on SNR or PSNR, we show the results of our networks for comparison. 
The underlying map is $y=f(\bm{x})=f(x,y)$ where $y$ represents a greyscale intensity, and $\bm{x}$ encodes the location of each pixel.
We use the signal-to-noise ratio (SNR) as the primary evaluation metric to measure reconstruction quality, where higher SNR indicates better accuracy. 
SNR and PSNR are defined in decibels (dB) using the expressions
\begin{equation}
    \text{SNR (dB)} = 10\,\log_{10}\left(\frac{P_{\text{signal}}}{\text{MSE}}\right),\,\,\text{and}\,\,
    \text{PSNR (dB)} = 10\,\log_{10}\left(\frac{\text{MAX}^{2}_{\text{signal}}}{\text{MSE}}\right),
\end{equation}
where $P_{\text{signal}}$ is the mean square image intensity,  $\text{MAX}^{2}_{\text{signal}}$ is the square of the maximum signal intensity and $\text{MSE}$ the mean square difference between the original and reconstructed image. 

\begin{figure}[!ht]
    \centering
    \includegraphics[width=\textwidth]{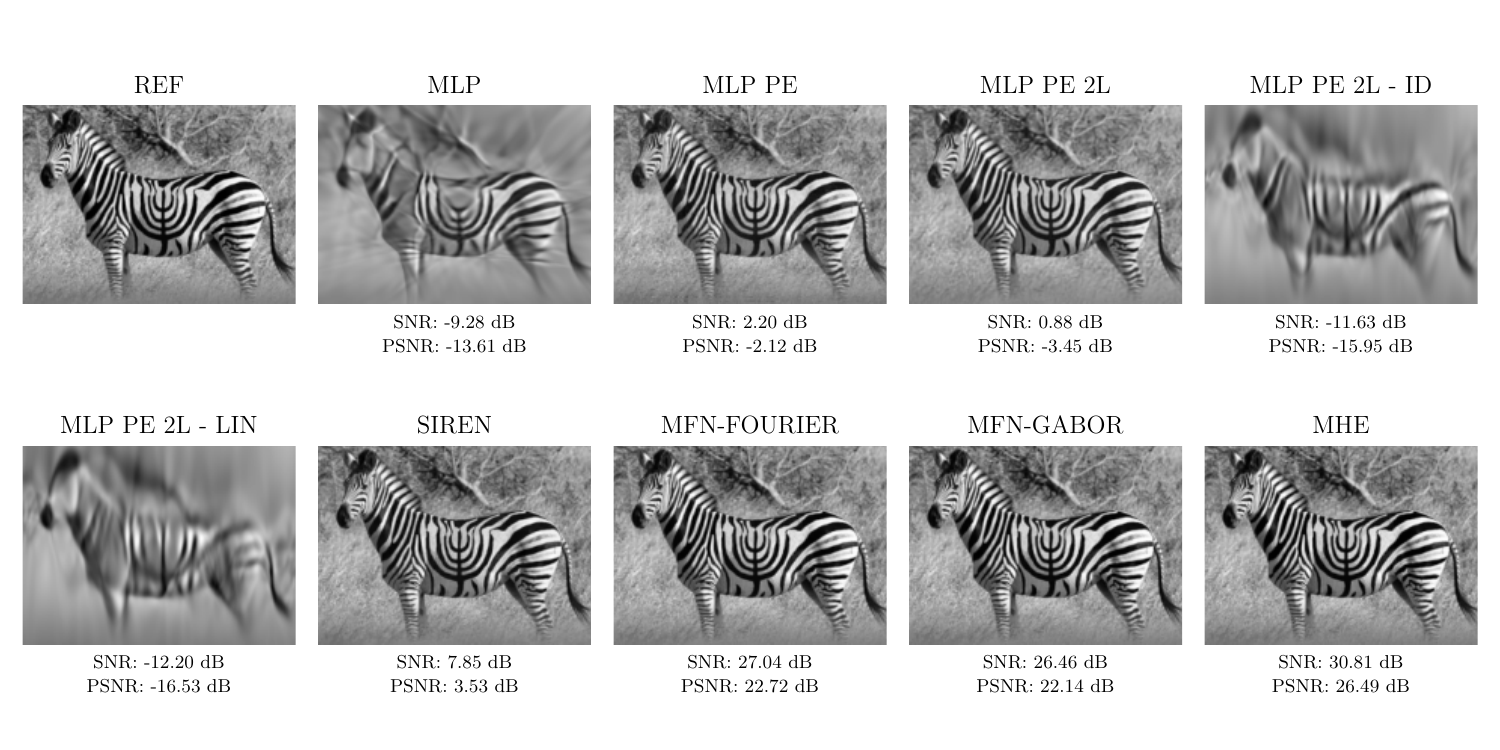}
    \caption{Resulting INR reconstructions - XY Zebra.}
    \label{fig:res_xy_zebra}
\end{figure}

Vanilla MLP and linear encoding are unable to mitigate the effect of spectral bias, and application of frequency-based position encoding becomes essential. 
Overall, MHE produced the best image reconstruction (SNR 30.81 dB), followed by MFN-Fourier (27.04 dB) and MFN-Gabor (26.46 dB), as shown in Figure~\ref{fig:res_xy_zebra}.

\subsection{XYZ Pipe}\label{sec:xyz_pipe}

We extend the reconstructions to three spatial variables where $\bm{x}=(x,y,z)\subset\mathbb{R}^3$, and evaluate the performance of our networks to learn pressures and velocities at the nodes of an unstructured mesh. 
We consider both spatially-varying \emph{scalar} fields, representing either the pressure or one of the three velocity components, and \emph{vector} (multi-component) fields, where the network outputs include all three velocity components, plus pressure.

\begin{figure}[!ht]
    \centering
    \begin{subfigure}[b]{0.48\textwidth}
        \centering
        \includegraphics[width=\textwidth]{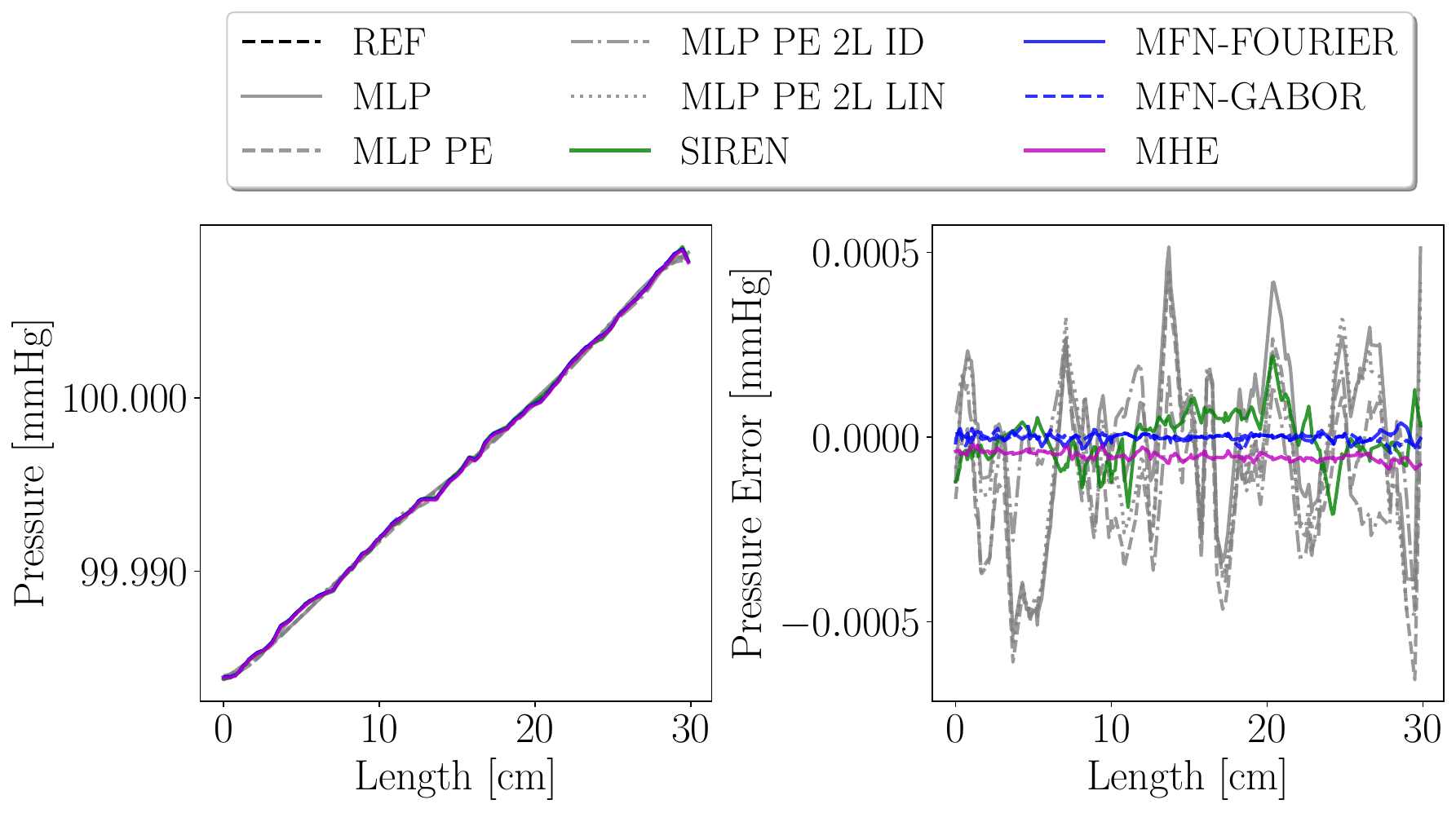}
        \caption{Single-field pressure reconstruction.}
        \label{fig:xyz_pipe_profiles_p}
    \end{subfigure}
    \hfill
    \begin{subfigure}[b]{0.48\textwidth}
        \centering
        \includegraphics[width=\textwidth]{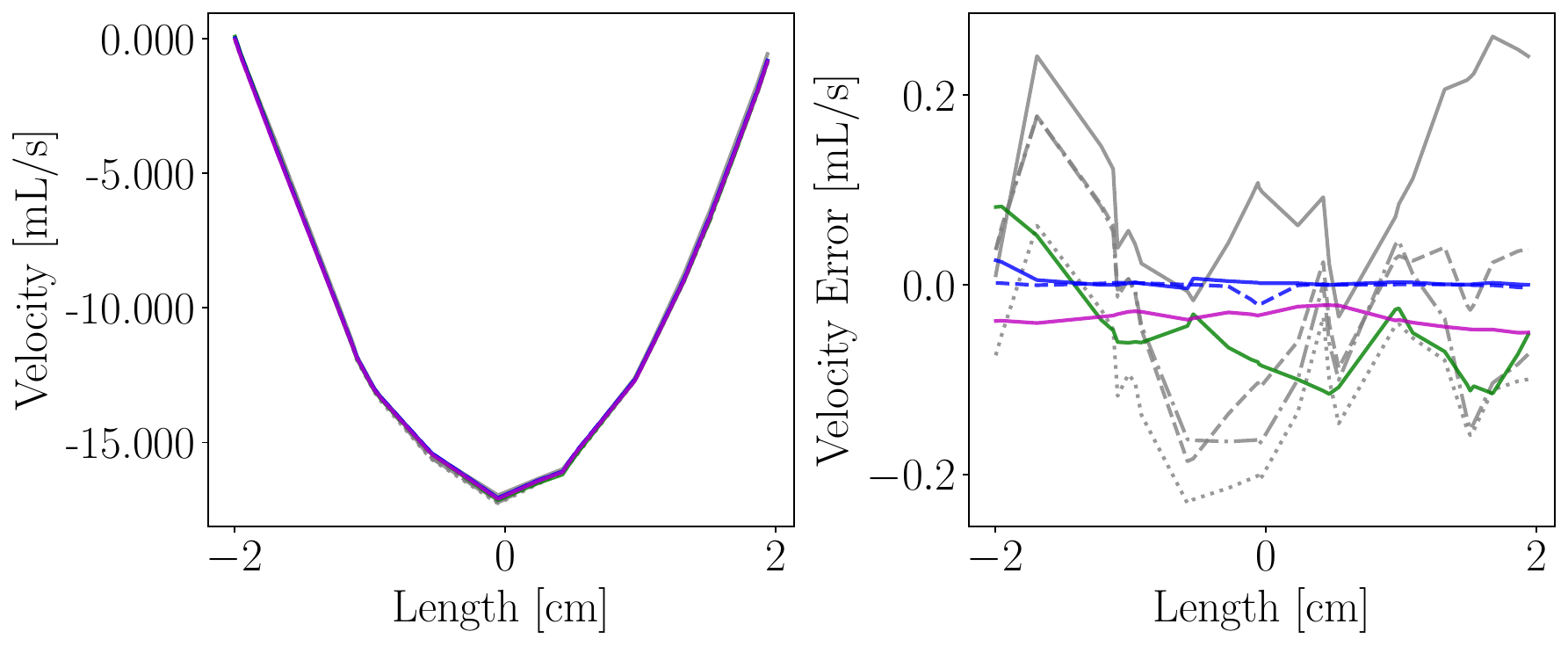}
        \caption{Single-field Z velocity reconstruction.}
        \label{fig:xyz_pipe_profiles_vz}    
    \end{subfigure}

    \begin{subfigure}[b]{0.48\textwidth}
        \centering
        \includegraphics[width=\textwidth]{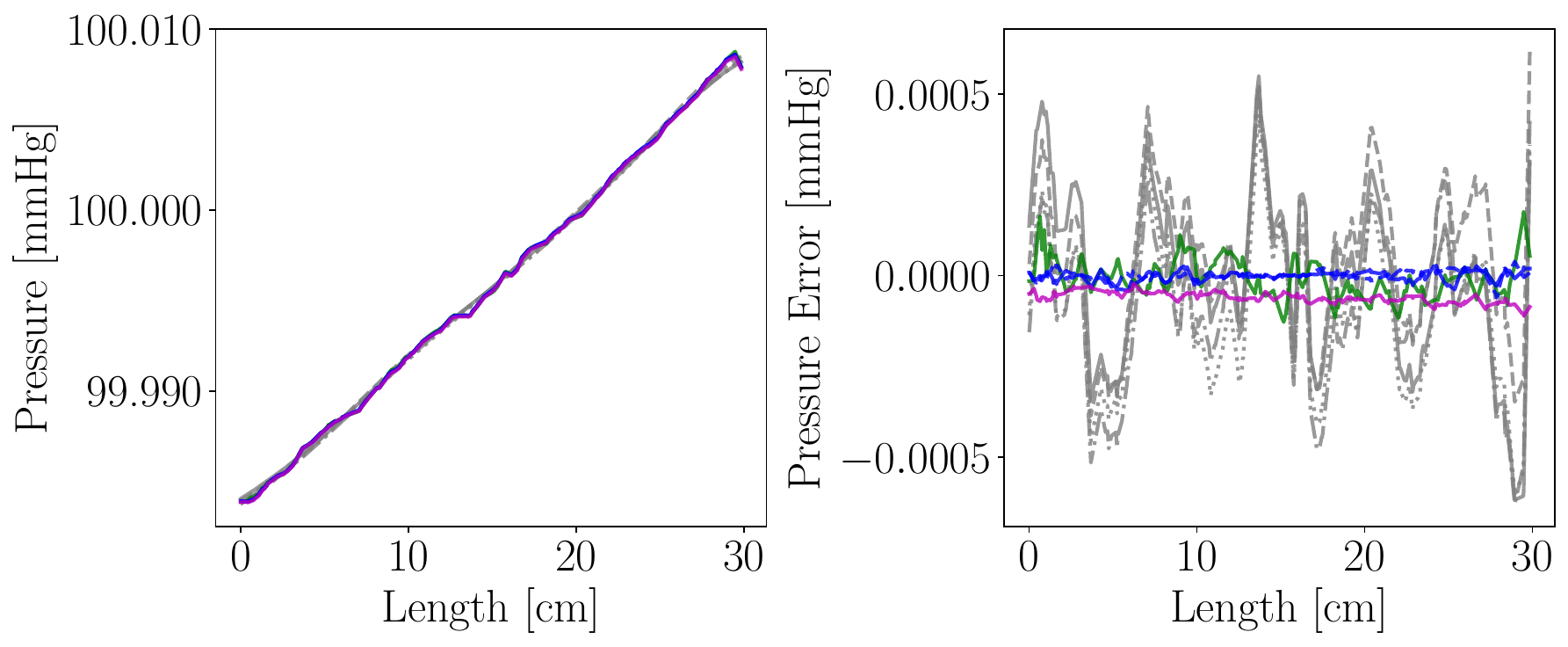}
        \caption{Multi-field pressure reconstruction.}
        \label{fig:xyz_pipe_profiles_pressure_vp}
    \end{subfigure}
    \hfill
    \begin{subfigure}[b]{0.48\textwidth}
        \centering
        \includegraphics[width=\textwidth]{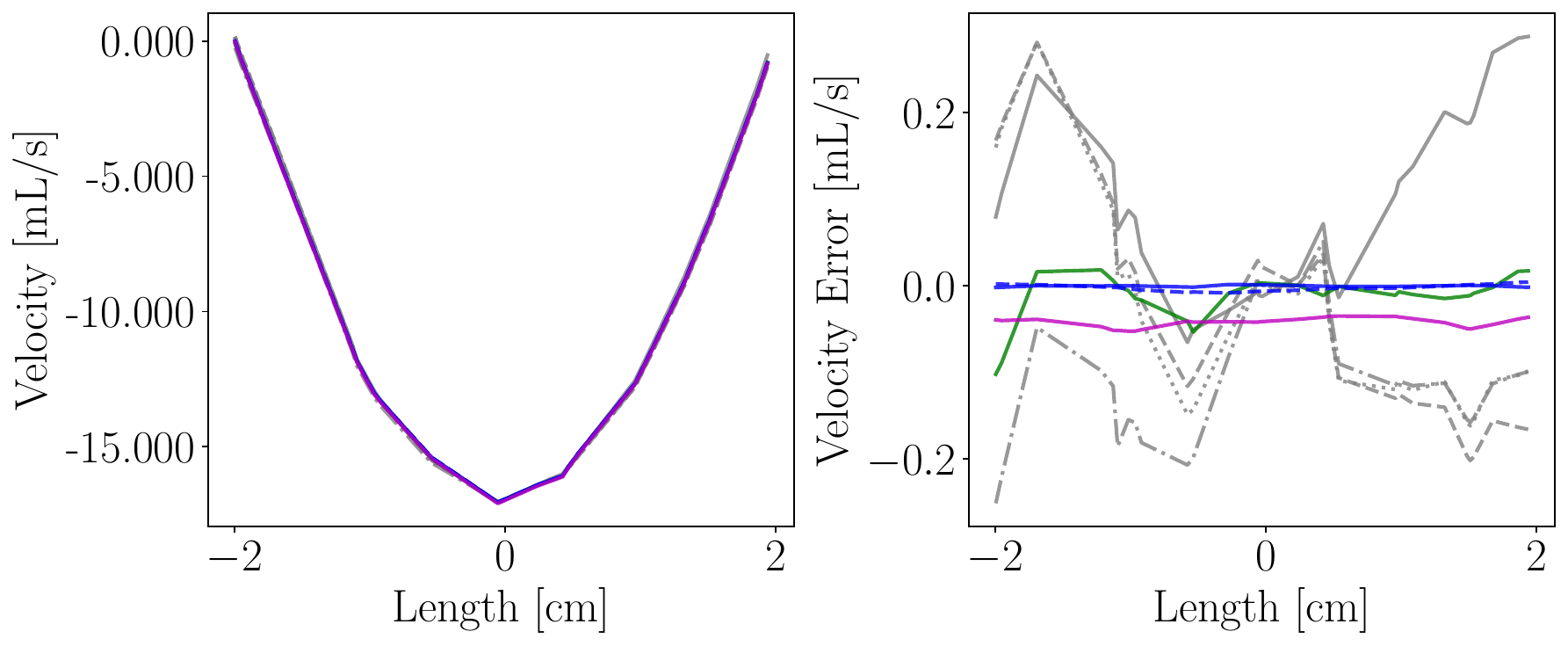}
        \caption{Multi-field Z velocity reconstruction.}
        \label{fig:xyz_pipe_profiles_vel_vp}    
    \end{subfigure}
    \caption{Single- and multi-fields reconstruction of pressure and axial velocity - XYZ Pipe.}
    \label{fig:xyz_pipe}
\end{figure}

All networks are able to reconstruct a steady pressure field with high accuracy as shown in Figure~\ref{fig:xyz_pipe} and Table~\ref{tab:xyz_pipe_rmse}. 
The reconstruction accuracy is generally greater for the pressure reconstruction due to the higher degree of smoothness (linear spatial profile) and lower for the velocity. Learning multiple components at the same time does not significantly affect reconstruction accuracy except for MFN-Fourier and MFN-Gabor where learning multiple components leads to better velocity reconstructions. 
Among the networks, MFN-Gabor shows the best performance, followed by MFN-Fourier and MHE.
We then used MFN-Gabor to understand how reconstruction accuracy is affected by number of iterations and batch size.
The results in Table~\ref{tab:xyz_pipe_rmse_sweep} confirm that accuracy generally increases with both the number of iterations and the batch size, and that simple hemodynamic result field are representable by the selected architectures to a high degree of accuracy. 

\begin{table}[!ht]
\centering
\caption{Single- and multi-component reconstruction errors - XYZ Pipe.}
\begin{tabular}{l c c | c c}
\toprule
\multirow{2}{*}{\bf Models} & \multicolumn{2}{c|}{\bf Single-Component} & \multicolumn{2}{c}{\bf Multi-Component}\\
             & RMSE $p$ & RMSE $v_{z}$ & RMSE $p$ & RMSE $v_{z}$\\
\midrule
MLP           & 3.886$\times$10$^{-4}$ & 2.340$\times$10$^{-1}$ & 3.838$\times$10$^{-4}$ & 2.343$\times$10$^{-1}$\\
MLP PE        & 3.799$\times$10$^{-4}$ & 1.531$\times$10$^{-1}$ & 3.842$\times$10$^{-4}$ & 2.126$\times$10$^{-1}$\\
MLP PE 2L ID  & 3.182$\times$10$^{-4}$ & 1.749$\times$10$^{-1}$ & 3.965$\times$10$^{-4}$ & 2.331$\times$10$^{-1}$\\
MLP PE 2L LIN & 3.816$\times$10$^{-4}$ & 1.808$\times$10$^{-1}$ & 3.862$\times$10$^{-4}$ & 1.801$\times$10$^{-1}$\\
SIREN         & 1.176$\times$10$^{-4}$ & 6.966$\times$10$^{-2}$ & 1.247$\times$10$^{-4}$ & 6.589$\times$10$^{-2}$\\
MFN-Fourier   & 1.562$\times$10$^{-5}$ & $\boldsymbol{1.162\times10^{-2}}$ & 1.592$\times$10$^{-5}$ & 5.909$\times$10$^{-3}$\\
MFN-Gabor     & $\boldsymbol{1.048\times10^{-5}}$ & 1.408$\times$10$^{-2}$ & $\boldsymbol{1.062\times10^{-5}}$ & $\boldsymbol{5.023\times10^{-3}}$\\
MHE           & 5.213$\times$10$^{-5}$ & 3.993$\times$10$^{-2}$ & 6.857$\times$10$^{-5}$ & 4.623$\times$10$^{-2}$\\
\bottomrule
\end{tabular}
\label{tab:xyz_pipe_rmse}
\end{table}

\begin{table}[!ht]
\centering
\caption{RMSE for MFN-Gabor following hyperparameter sweep - XYZ Pipe.}
\begin{tabular}{c c c c c}
\toprule
\multirow{2}{*}{\bf It./Batch Size} & \multicolumn{2}{c}{\bf Reconstructing $(v_x,v_y,v_z,p)$} & \multicolumn{2}{c}{\bf Reconstructing $(v_z,p)$}\\
             & RMSE $p$ & RMSE $v_{z}$ & RMSE $p$ & RMSE $v_{z}$\\
\midrule
10K/1024 & 1.093$\times$10$^{-5}$ & 4.803$\times$10$^{-3}$ & 1.005$\times$10$^{-5}$ & 4.648$\times$10$^{-3}$\\
10K/2048 & 1.089$\times$10$^{-5}$ & 5.386$\times$10$^{-3}$ & 8.483$\times$10$^{-6}$ & 3.204$\times$10$^{-3}$\\
10K/4096 & 8.179$\times$10$^{-6}$ & 3.013$\times$10$^{-3}$ & 8.002$\times$10$^{-6}$ & 2.898$\times$10$^{-3}$\\
\midrule
20K/1024 & 1.214$\times$10$^{-5}$ & 6.138$\times$10$^{-3}$ & 1.049$\times$10$^{-5}$ & 5.690$\times$10$^{-3}$\\
20K/2048 & 1.095$\times$10$^{-5}$ & 4.925$\times$10$^{-3}$ & 8.512$\times$10$^{-6}$ & 4.022$\times$10$^{-3}$\\
20K/4096 & 8.432$\times$10$^{-6}$ & 3.091$\times$10$^{-3}$ & 8.278$\times$10$^{-6}$ & 3.162$\times$10$^{-3}$\\
\midrule
30K/1024 & 1.219$\times$10$^{-5}$ & 6.982$\times$10$^{-3}$ & 1.143$\times$10$^{-5}$ & 6.141$\times$10$^{-3}$\\
30K/2048 & 8.997$\times$10$^{-6}$ & 4.577$\times$10$^{-3}$ & 8.412$\times$10$^{-6}$ & 3.005$\times$10$^{-3}$\\
30K/4096 & 8.997$\times$10$^{-6}$ & 4.577$\times$10$^{-3}$ & 8.412$\times$10$^{-6}$ & 3.005$\times$10$^{-3}$\\
\bottomrule
\end{tabular}
\label{tab:xyz_pipe_rmse_sweep}
\end{table}

\subsection{XYZ Aorta}\label{sec:xyz_aorta}

\begin{figure}[!ht]
\centering
\begin{subfigure}{0.48\textwidth}
\centering
\includegraphics[width=\textwidth]{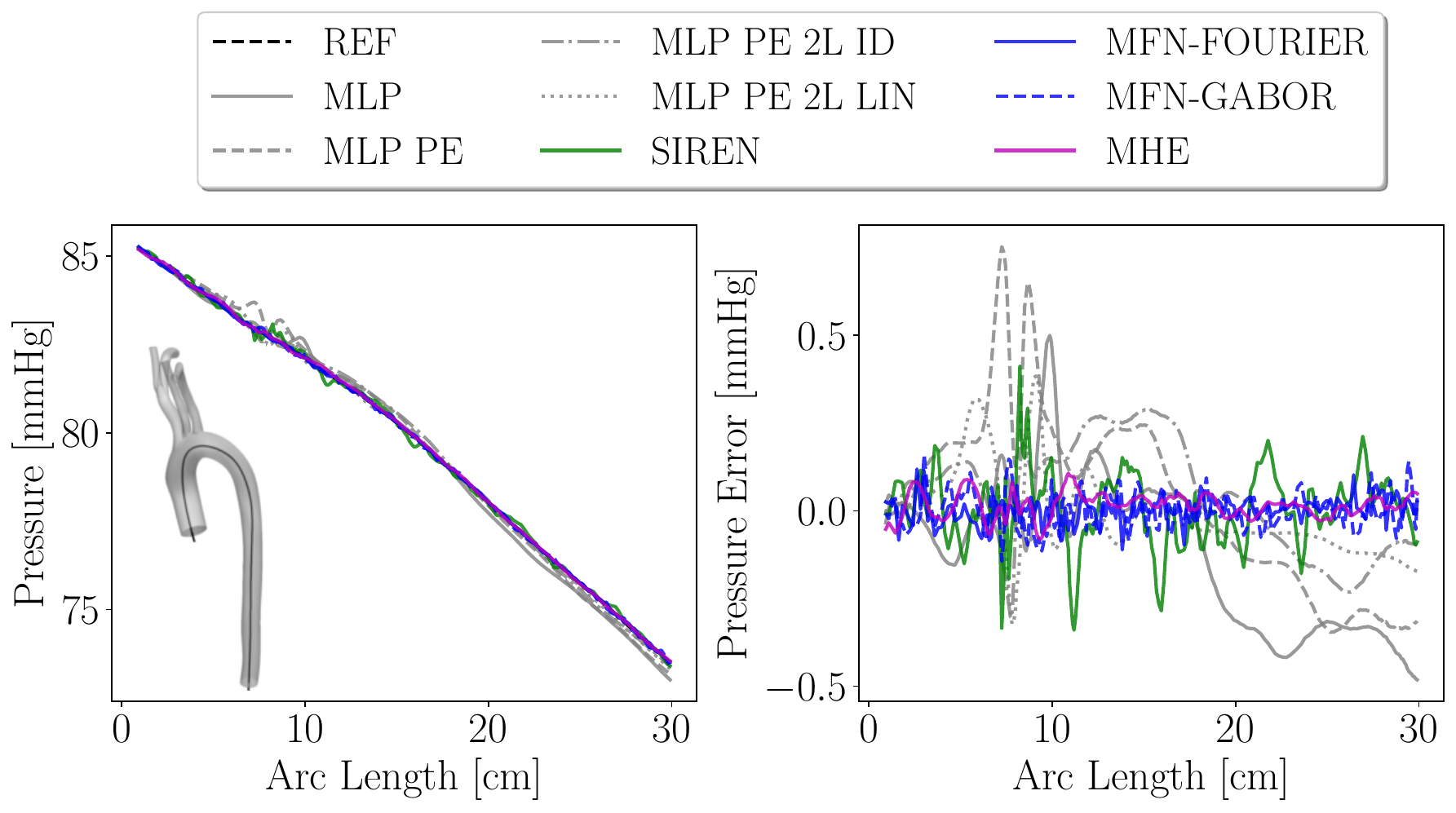}
\caption{Pressure profile and error.}
\label{fig:xyz_aorta_uscale_profile_p}
\end{subfigure}
\begin{subfigure}{0.48\textwidth} 
\centering
\includegraphics[width=\textwidth]{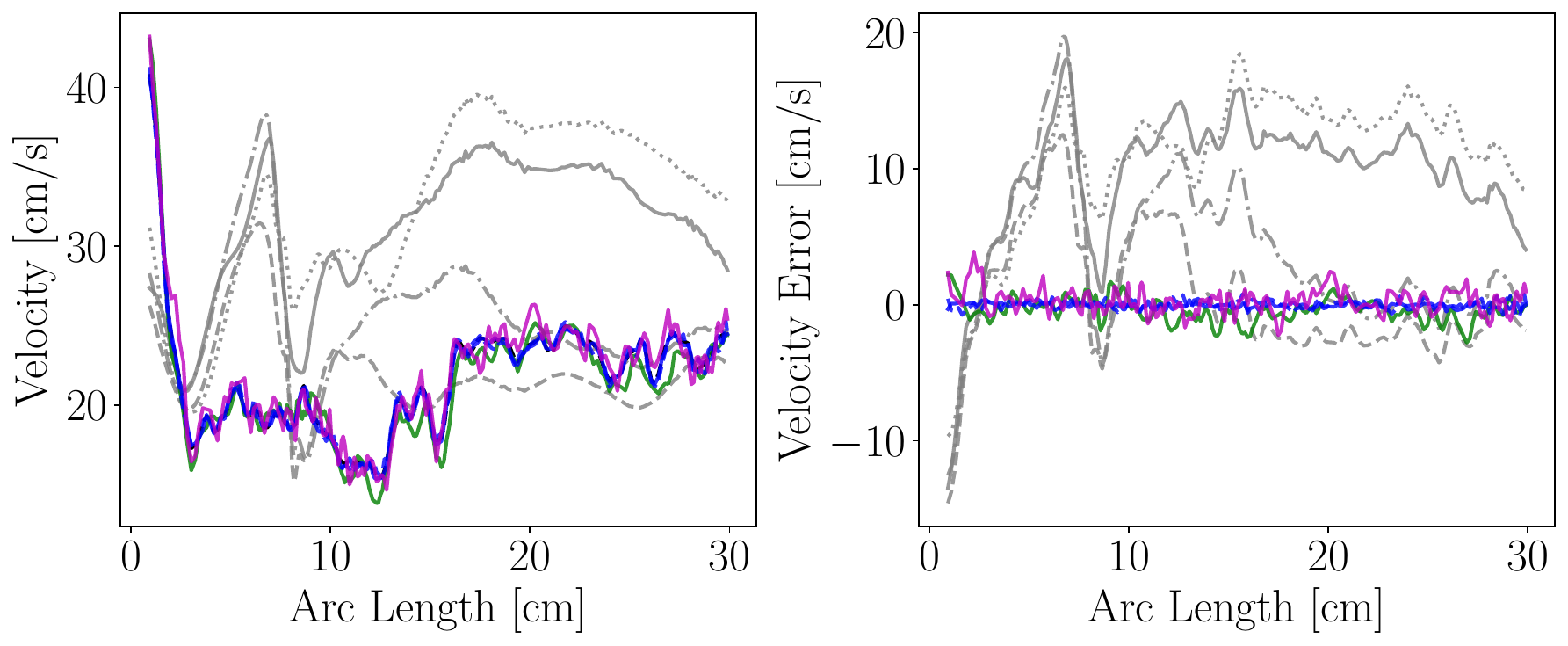}
\caption{Velocity norm profile and error.}
\label{fig:xyz_aorta_uscale_profile_vnorm}
\end{subfigure}
\caption{Pressure/velocity reconstructions and associated errors - XYZ Aorta.}\label{fig:xyz_aorta_profiles_rmse}
\end{figure}

\begin{figure}[!ht]
\centering
\begin{subfigure}[c]{0.45\textwidth}
\centering
\includegraphics[width=\textwidth]{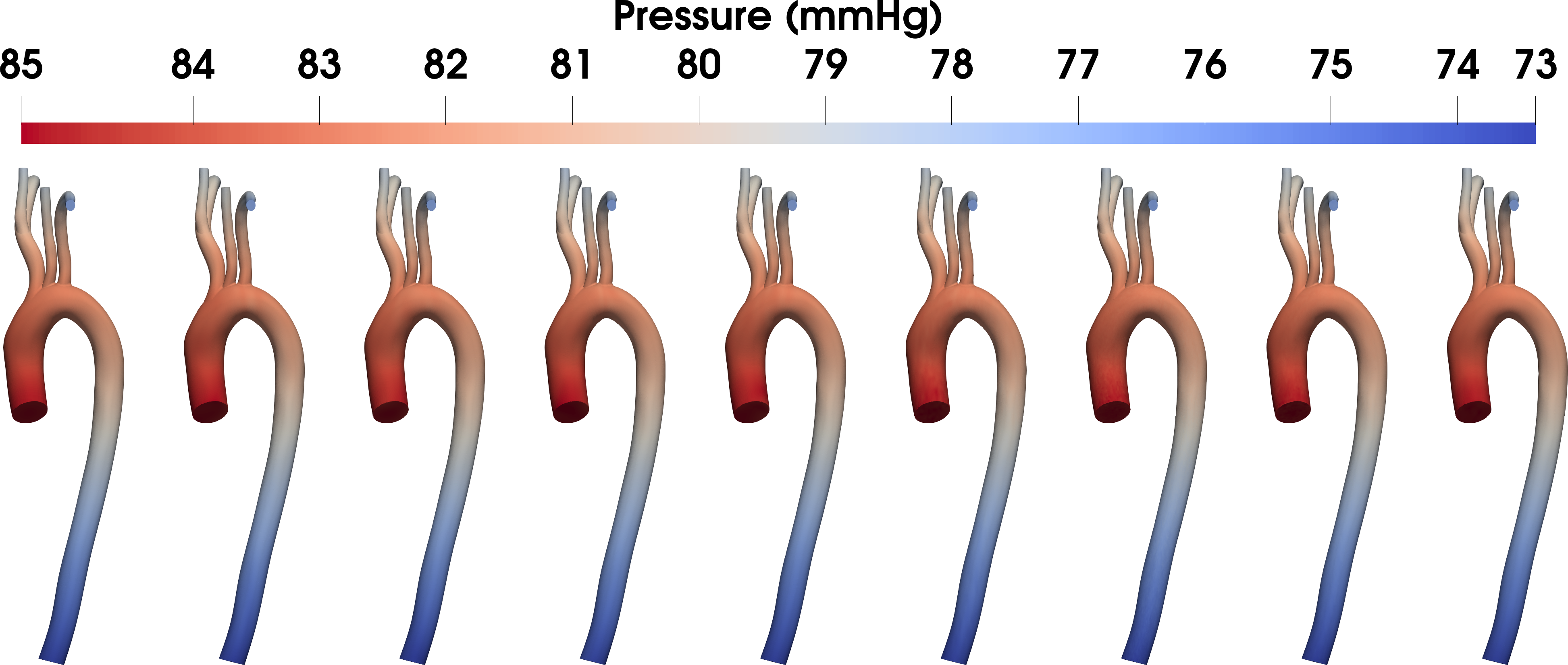}
\caption{From left to right: Ref., MLP, MLP PE, MLP PE 2L ID, MLP PE 2L LIN, SIREN, MFN-Fourier, MFN-Gabor, MHE.}
\label{fig:xyz_aorta_rec_pressure}
\end{subfigure}
$\quad$
\begin{subfigure}[c]{0.46\textwidth}
\centering
\includegraphics[width=\textwidth]{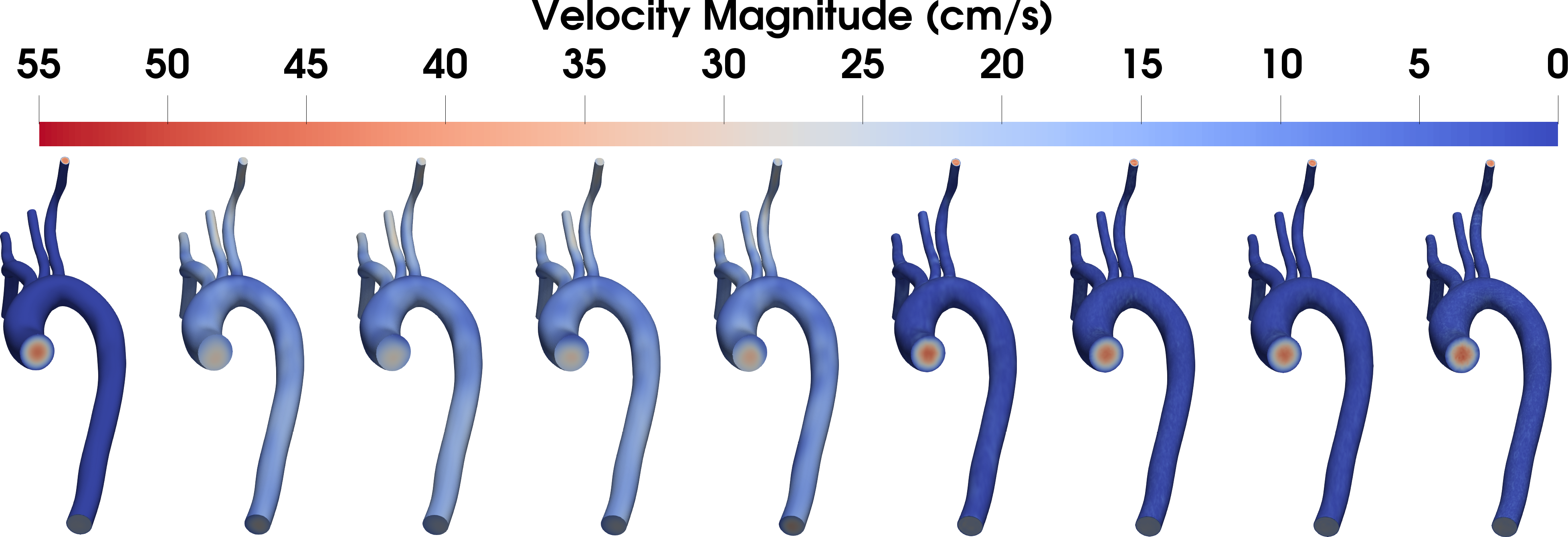}
\caption{From left to right: Ref., MLP, MLP PE, MLP PE 2L ID, MLP PE 2L LIN, SIREN, MFN-Fourier, MFN-Gabor, MHE.}
\label{fig:xyz_aorta_rec_vnorm}
\end{subfigure}
\caption{Pressure and velocity reconstructions contours - XYZ Aorta.}\label{fig:xyz_aorta_rec}
\end{figure}

We consider the reconstruction of pressure and velocity fields from the steady state simulation of a patient-specific aortic anatomy.
Figure~\ref{fig:xyz_aorta_profiles_rmse} and Table~\ref{tab:xyz_aorta_rmse} suggest how the pressure and velocity field complexity significantly increase with respect to the steady state solution of an idealized cylindrical domain.
The results also show how more sophisticated approaches (SIREN, MFNs and MHE) offer superior performance with respect to simple MLP architectures with frequency based positional encoding, particularly when capturing the velocity field, which shows a reduced degree of regularity with respect to the pressure field. 
In addition, Figure~\ref{fig:xyz_aorta_rec} and Table~\ref{tab:xyz_aorta_rmse} confirm that the reconstruction error offered by SIREN, MFN and NHE is negligible for the pressure and approximately in the order of 1 cm/s for the velocity, with reference centerline steady state velocities in the order of 15-20 cm/s.
The error further reduces by increasing the number of iterations and batch size at a sublinear rate. 
Additionally, velocity errors computed by MLP with identity or linear positional encoding are much larger than pressure errors for the XYZ Aorta test case, given the reduced smoothness in the velocities.
Finally, we discuss the compression offered by neural representations. The size of the checkpoint file for the networks ranges from 4.4MB for MLP to 71.6MB for MHE under baseline parameters (see Section~\ref{sec:hyper_choice}). Since the network operates on single-precision floating point arithmetic and is designed to predict four scalar components, the memory required to store four floats at each node in the mesh is 554,474$\cdot$4$\cdot$4/1024$^{2}$ = 8.5 MB, where we do not consider the memory needed to store the mesh coordinates and element connectivity, since these are also used by the networks to reconstruct the field through shape function interpolation. For steady state results of mesh with approximately 3M finite element, the compression ratio therefore ranges from $\sim$0.12 to $\sim$1.9, where for MHE, the size of the encoding dominates over the size of the network. Therefore, for steady state simulation results there is no clear advantange in using neural representations. 

\begin{table}[!ht]
\caption{RMSE results for XYZ Aorta.}
\begin{subtable}[c]{0.45\textwidth}
\centering
\caption{Errors in pressure and velocity norm.}
\begin{tabular}{l c c c}
\toprule
{\bf Models} & {\bf RMSE $p$} & {\bf RMSE $\Vert \bm{v}\Vert_{2}$}\\
\midrule
MLP           & 1.517$\times$10$^{-1}$ & 7.817\\
MLP PE        & 1.041$\times$10$^{-1}$ & 7.983\\
MLP PE 2L ID  & 1.600$\times$10$^{-1}$ & 7.814\\
MLP PE 2L LIN & 9.062$\times$10$^{-2}$ & 7.355\\
SIREN         & 6.969$\times$10$^{-2}$ & 1.517\\
MFN-Fourier   & 4.785$\times$10$^{-2}$ & 1.155\\
MFN-Gabor     & $\boldsymbol{3.161\times10^{-2}}$ & $\boldsymbol{9.887\times10^{-1}}$\\
MHE           & 3.935$\times$10$^{-2}$ & 1.402\\
\bottomrule
\end{tabular}
\label{tab:xyz_aorta_rmse}
\end{subtable}\hfill
\begin{subtable}[c]{0.45\textwidth}
\centering
\caption{RMSE for MFN-Gabor from hyperparameter sweep.}
\begin{tabular}{c c c c c}
\toprule
{\bf It./Batch Size} & {\bf RMSE $p$} & {\bf RMSE $\Vert\bm{v}\Vert_{2}$}\\
\midrule
10K/1024 & 3.297$\times$10$^{-2}$ & 1.046\\
10K/2048 & 2.741$\times$10$^{-2}$ & 8.088$\times$10$^{-1}$\\
10K/4096 & 2.045$\times$10$^{-2}$ & 6.328$\times$10$^{-1}$\\
\midrule
20K/1024 & 2.270$\times$10$^{-2}$ & 7.295$\times$10$^{-1}$\\
20K/2048 & 1.869$\times$10$^{-2}$ & 5.900$\times$10$^{-1}$\\
20K/4096 & 1.547$\times$10$^{-2}$ & 4.543$\times$10$^{-1}$\\
\midrule
30K/1024 & 1.935$\times$10$^{-2}$ & 5.572$\times$10$^{-1}$\\
30K/2048 & 1.536$\times$10$^{-2}$ & 4.740$\times$10$^{-1}$\\
30K/4096 & 1.536$\times$10$^{-2}$ & 4.740$\times$10$^{-1}$\\
\bottomrule
\end{tabular}
\label{tab:xyz_aorta_gabor_rmse_sweep}
\end{subtable}
\end{table}

\subsection{XYZT Pipe}\label{sec:xyzt_pipe}
\begin{figure}[!ht]
\centering
\begin{subfigure}[b]{\textwidth}
\centering
\includegraphics[width=\textwidth]{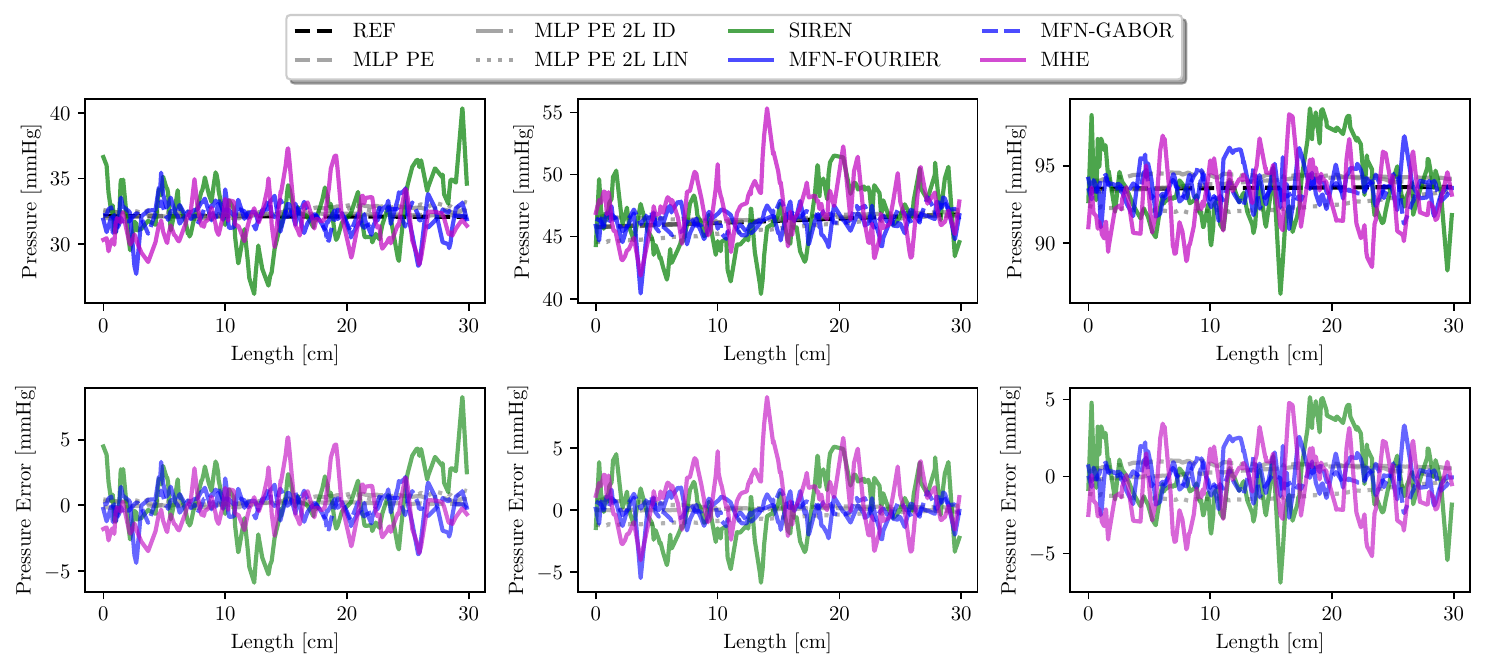}
\caption{Reconstructed pressure profile along cylinder axis for time index 40 (left), 80 (center), and 120.}
\label{fig:xyzt_pipe_profile_p_space}
\end{subfigure}
\hfill
\begin{subfigure}[b]{\textwidth} 
\centering
\includegraphics[width=\textwidth]{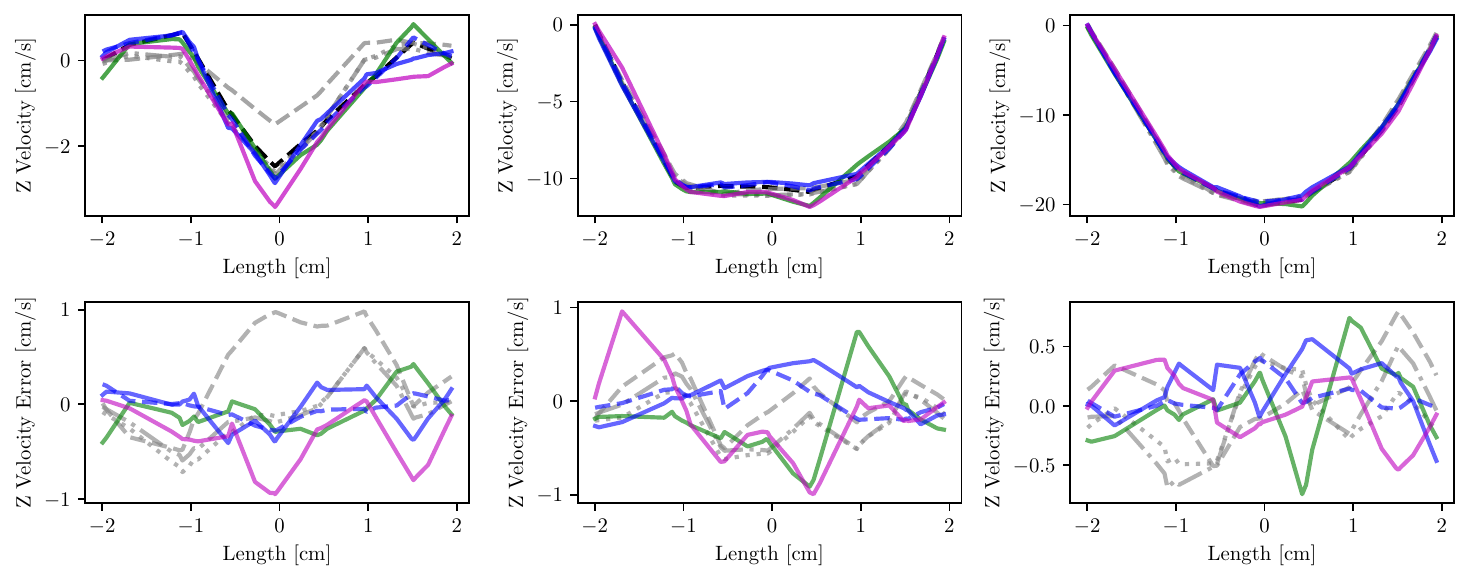}
\caption{Reconstructed Z velocity profile orthogonal to the cylinder axis for a time index that corresponds to the minimum (left), maximum (right), and the point halfway between minimum and maximum of the sinusoidal response in time.}
\label{fig:xyzt_pipe_profile_vz_space}
\end{subfigure}    
\caption{Pressure and velocity spatial profiles at different times - XYZT Pipe.}
\label{fig:xyzt_pipe_profile_space}
\end{figure}

\begin{figure}[!ht]
\centering
\begin{subfigure}[b]{0.48\textwidth}
\centering
\includegraphics[width=\textwidth]{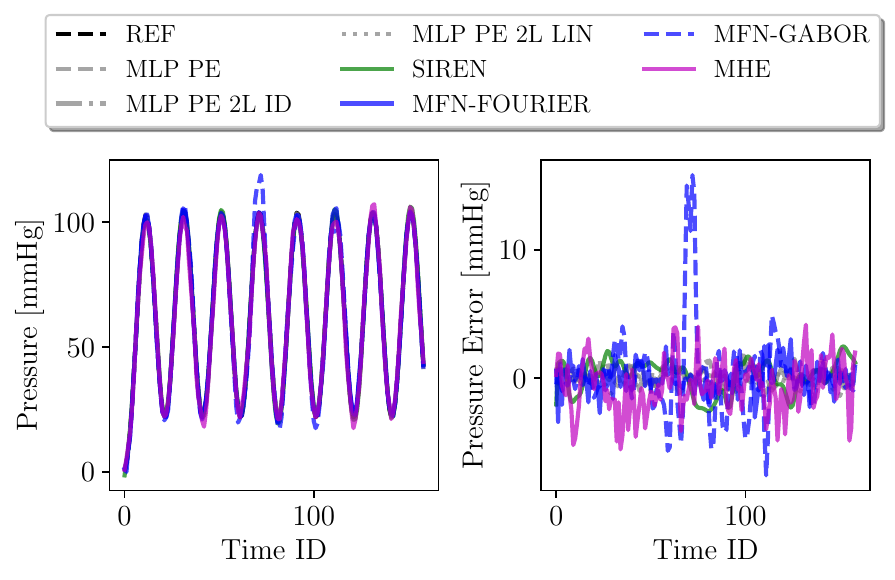}
\caption{Pressure versus time (space location ID 30).}
\label{fig:xyzt_pipe_profile_p_time}
\end{subfigure}
\hfill
\begin{subfigure}[b]{0.48\textwidth} 
\centering
\includegraphics[width=\textwidth]{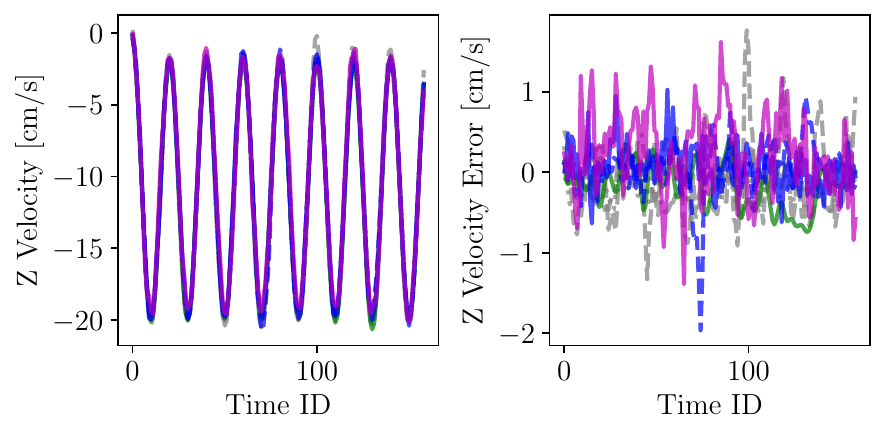}
\caption{Z Velocity versus time (space location ID 30).}
\label{fig:xyzt_pipe_profile_vz_time}
\end{subfigure}    
\caption{Pressure and velocity temporal profiles at a fixed spatial location - XYZT Pipe.}
\label{fig:xyzt_pipe_profile_time}
\end{figure}

%
We now focus on the representations of time- and space-dependent fields.
Due to the similarities to the XT Pipe test case in Section~\ref{sec:xt_pipe}, we expect similar accuracies. However the results are different due to two main reasons. 
First, the number of times nodal values are seen at training is significantly different for the two test cases. The XT Pipe dataset contains 1000 space locations and 60 time steps. Since the training consisted in 10,000 iterations with batch 1024, the entire dataset was seen $\sim$170 times during training. 
Conversely, for XYZT Pipe the dataset consists in 2,354 nodes over 160 time steps, i.e., it was seen $\sim$27 times. Second, while a single pressure component is trained in XT Pipe, four components with varying smoothness were simultaneously learned for XYZT Pipe.
Resulting RMSE accuracies as shown in Table~\ref{tab:xyzt_pipe_rmse}.
Best results for pressures are obtained from fixed frequency encoding in time and identity and linear space encoding, where advanced encoding (SIREN, MFN and MHE) provide worse performance. By increasing the number of iterations to 60,000 and training only on the pressure, we obtained results in line with those reported for the XT Pipe test case in Table~\ref{tab:xt_pipe_rmse}.
In addition, MFN-Fourier provides the best velocity reconstructions, that however appear similar for all approaches.
Spatial velocity profiles in Figure~\ref{fig:xyzt_pipe_profile_space} show errors of approximately 1.0 cm/s in diastole and half the same errors at systole, due to a higher SNR.
Similarly, pressure oscillations as high as $\sim$5 mmHg can be observed for SIREN and MHE. 
Additionally, Figure~\ref{fig:xyzt_pipe_profile_time} shows that all networks are able to accurately capture the oscillatory system response in time, with overshooting only visible for MFN-Fourier in one of the cycles. 
Since we use the same networks for all test cases, their size ranges from 4.4MB of an MLP network to 71.6MB for MHE.
To represent a fully pulsatile simulation, a mesh representation would instead require 2,354$\cdot$4$\cdot$4$\cdot$160/1024$^{2}$=5.7MB, resulting in no significant compression, similar to our conclusion in the previous section for steady-state results.

\begin{table}[!ht]
\caption{RMSE results - XYZT Pipe.}
\label{tab:xyzt_pipe_rmse}
\begin{subtable}[c]{0.45\textwidth}
\centering
\caption{Training for 10K iterations on $p$ and $\bm{v}$.}
\label{tab:xyzt_pipe_rmse_10K}
\begin{tabular}{l c c}
\toprule
\textbf{Model} & \textbf{RMSE} $p$ & \textbf{RMSE} $v_{z}$\\
\midrule
MLP           & 2.970$\times$10$^{+1}$ & 3.880\\
MLP PE        & 7.568$\times$10$^{-1}$ & 4.326$\times$10$^{-1}$\\
MLP PE 2L ID  & $\boldsymbol{6.341\times10^{-1}}$ & 3.670$\times$10$^{-1}$\\
MLP PE 2L LIN & 6.451$\times$10$^{-1}$ & 3.447$\times$10$^{-1}$\\
SIREN         & 2.729 & 3.919$\times$10$^{-1}$\\
MFN-Fourier   & 1.331 & $\boldsymbol{2.211\times10^{-1}}$\\
MFN-Gabor     & 1.706 & 3.800$\times$10$^{-1}$\\
MHE           & 2.235 & 3.995$\times$10$^{-1}$\\
\bottomrule
\end{tabular}
\end{subtable}\hfill
\begin{subtable}[c]{0.45\textwidth}
\centering
\caption{Training for 60K iterations only on $p$.}
\label{tab:xyzt_pipe_rmse_60K}
\begin{tabular}{l c}
\toprule
\textbf{Model} & \textbf{RMSE} $p$\\
\midrule
MLP           & 4.091$\times$10$^{-1}$\\
MLP PE        & 2.997$\times$10$^{-1}$\\
MLP PE 2L ID  & $\boldsymbol{1.489\times10^{-1}}$\\
MLP PE 2L LIN & 1.678$\times$10$^{-1}$\\
SIREN         & 1.371\\
MFN-Fourier   & 5.807$\times$10$^{-1}$\\
MFN-Gabor     & 4.932$\times$10$^{-1}$\\
MHE           & 8.398$\times$10$^{-1}$\\
\bottomrule
\end{tabular}
\end{subtable}
\end{table}

\subsection{The tradeoff between overfitting and interpolation}\label{sec:overfit_vs_interp}

In all test cases above, we focus on ability of a network to \emph{memorize} the values of a vector field at a number of fixed locations in space and time, i.e., the nodal solutions of a finite element simulation. As discussed in Section~\ref{sec:interp_memoriz} PE, SIREN, MFN and MHE mitigate spectral bias by promoting overfitting. This, however, leads to a sharp reduction in the abilities of the networks to approximate smooth fields, and justifies the use of shape function interpolation based on an underlying tetrahedral mesh.

In this section we focus on the generalization performance of the selected architectures, or their ability to produce accurate reconstructions over space-time locations unseen at training, particularly in the interior of elements. We quantify this ability using a regular validation grid, as shown in Figure~\ref{fig:xyzt_pipe_val_grids}.
Outside the nodal locations, the result accuracy looks very different except under limited overfitting offered by MLP-based architecture. 

\begin{table}[ht]
\begin{minipage}[b]{0.56\linewidth}
\centering
\begin{tabular}{l c c}
\toprule
\textbf{Model} & \textbf{RMSE} $p$ & \textbf{RMSE} $v_{z}$\\
\midrule
MLP           & 2.970$\times$10$^{+1}$ & $\boldsymbol{7.722}$\\
MLP PE        & 7.529$\times$10$^{-1}$ & 8.757\\
MLP PE 2L ID  & 6.203$\times$10$^{-1}$ & 8.845\\
MLP PE 2L LIN & $\boldsymbol{5.938\times10^{-1}}$ & 8.801\\
SIREN         & 1.677$\times$10$^{+1}$ & 8.304\\
MFN-Fourier   & 2.899$\times$10$^{+1}$ & 8.712\\
MFN-Gabor     & 3.163$\times$10$^{+1}$ & 8.508\\
MHE           & 2.537$\times$10$^{+1}$ & 8.842\\
\bottomrule
\end{tabular}
\caption{RMSE results over regular validation grid - XYZT Pipe.}
\label{table:student}
\end{minipage}\hfill
\begin{minipage}[b]{0.4\linewidth}
\centering
\includegraphics[width=\textwidth]{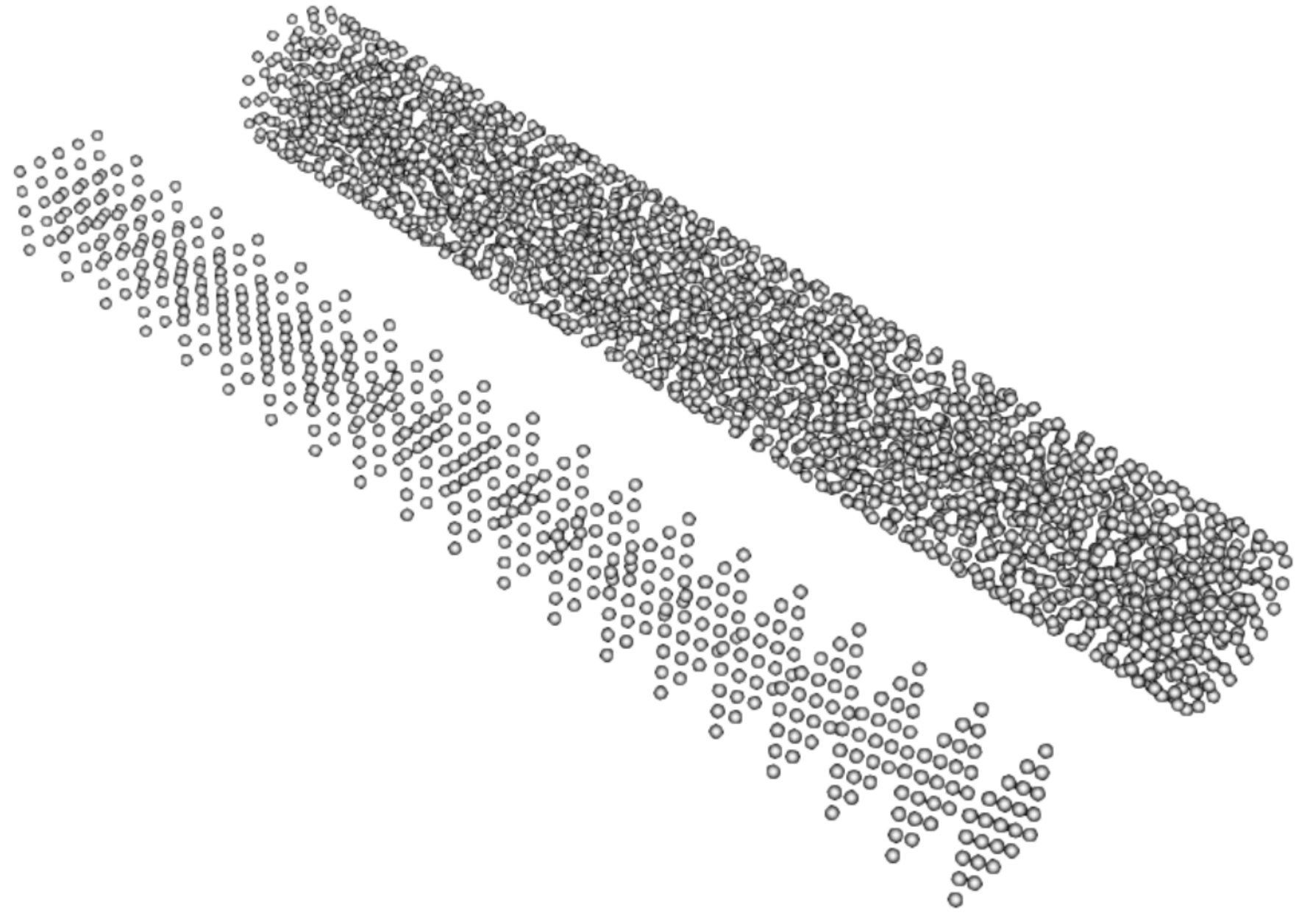}
\captionof{figure}{Unstructured training and regular validation grids.}
\label{fig:xyzt_pipe_val_grids}
\end{minipage}
\end{table}

\subsection{XYZT Aorta}\label{sec:xyzt_aorta}
%
Finally, we focus on learning a fully pulsatile hemodynamic field for a patient-specific aortic model containing 2,989,185 elements, 554,474 nodes and 180 time steps, which are representative of a typical cardiovascular simulation setup.
For this model we consider a range of training conditions from 10,000 iterations with 1,024 batch size, to 200,000 iterations with 4,096 batch size. This corresponds to see the entire dataset a number of times ranging from 1/10 of the total dataset to approximately 8 times. Therefore, in light of the relatively small number of times the networks see the entire dataset (which consist of 4 scalar components, one pressure and three velocities), the resulting accuracy is remarkable.
The best result was produced by the SIREN, followed by positional encoding with fixed frequencies and identity spatial encoding. MFN-Fourier consistently produced the worst results. 

\begin{figure}[!ht]
\centering
\begin{subfigure}[b]{\textwidth}
\centering
\includegraphics[width=\textwidth]{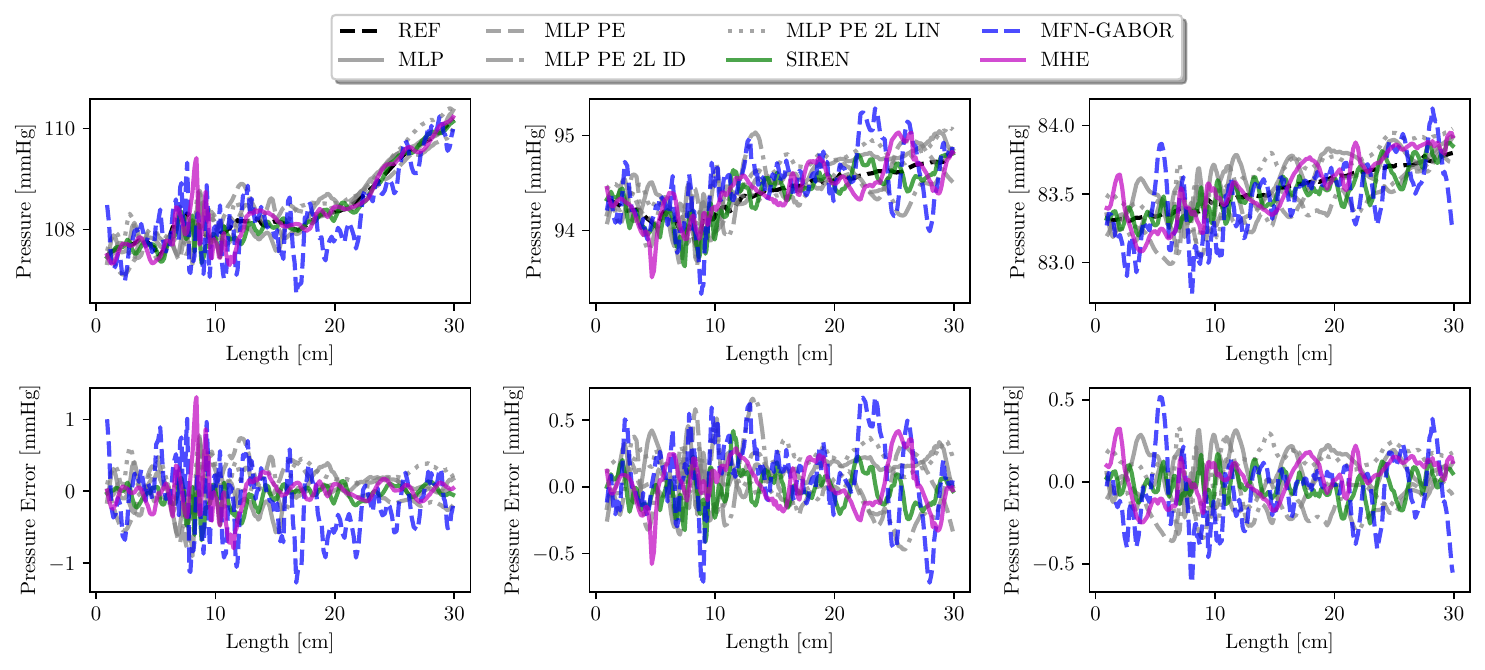}
\caption{Reconstructed pressure profile along aorta centerline path for time index 40 (left), 80 (center), and 120.}
\label{fig:xyzt_aorta_profile_p_space}
\end{subfigure}
\hfill
\begin{subfigure}[b]{\textwidth} 
\centering
\includegraphics[width=\textwidth]{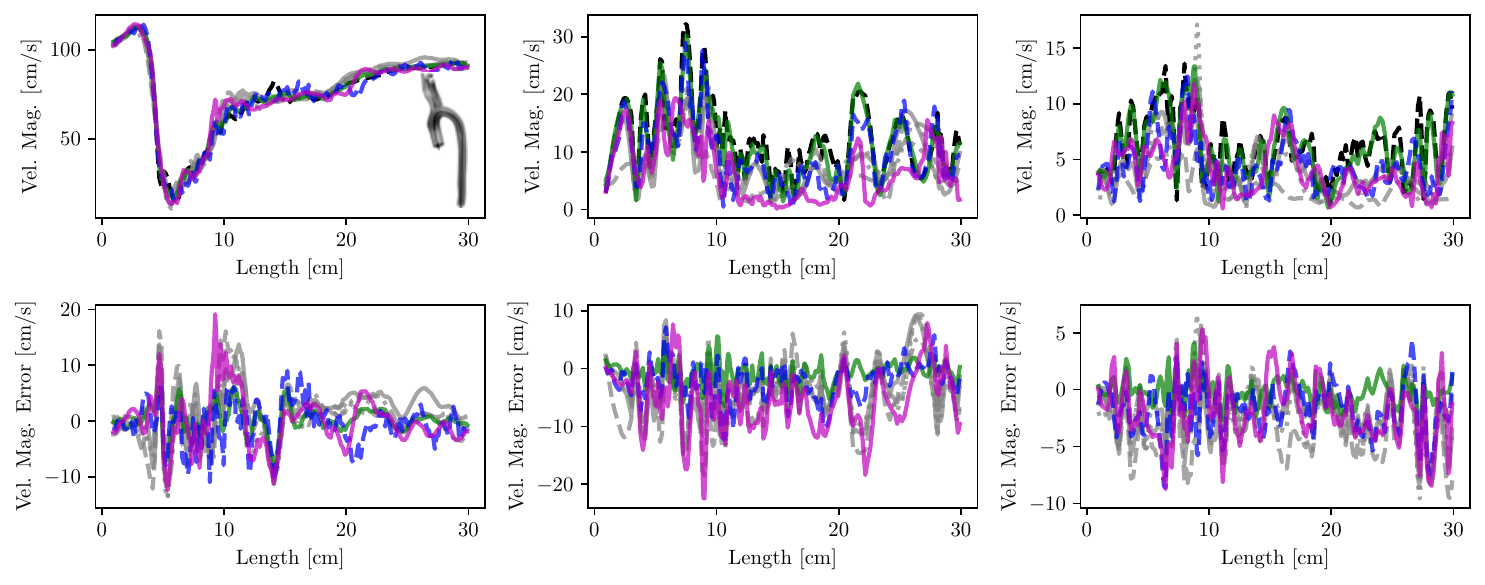}
\caption{Reconstructed velocity magnitude profile along aorta centerline path for time index 40 (left), 80 (center), and 120.}
\label{fig:xyzt_aorta_profile_vz_space}
\end{subfigure}    
\caption{XYZT Aorta test case. Pressure and velocity profiles along aorta centerline path at different times. These profiles are obtained after training a neural field representation for 200,000 iterations with 4096 batch size.}
\label{fig:xyzt_aorta_profiles_space}
\end{figure}

\begin{figure}[!ht]
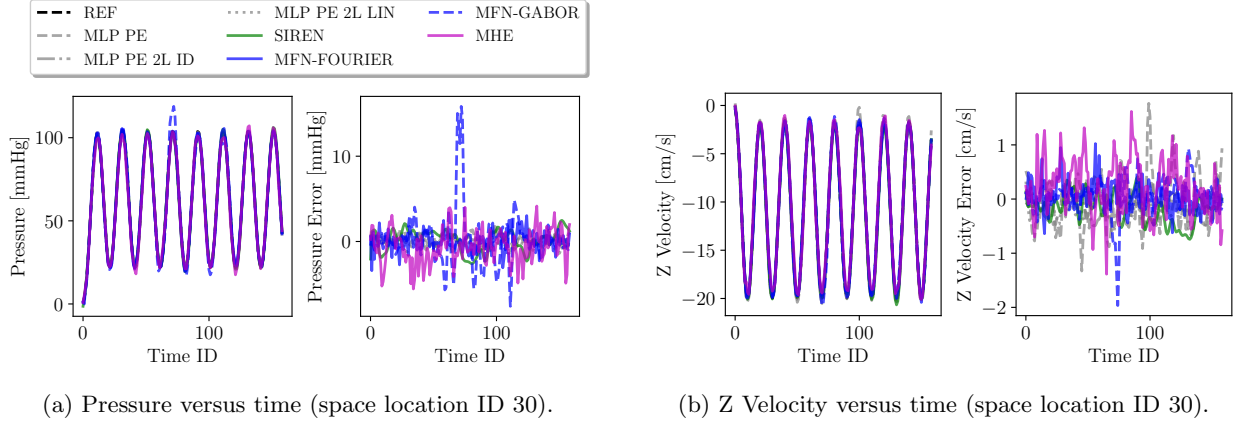

\centering
\begin{subfigure}[b]{0.48\textwidth}
\centering
\includegraphics[width=\textwidth]{fig_17a.pdf}
\caption{Pressure versus time (space location ID 30).}
\label{fig:xyzt_aorta_profile_p_time}
\end{subfigure}
\hfill
\begin{subfigure}[b]{0.48\textwidth} 
\centering
\includegraphics[width=\textwidth]{fig_17b.pdf}
\caption{Z Velocity versus time (space location ID 30).}
\label{fig:xyzt_aorta_profile_vz_time}
\end{subfigure}    
\caption{Pressure and velocity temporal profiles at a fixed spatial location - XYZT Aorta dataset.}
\label{fig:xyzt_aorta_profile_time}
\end{figure}

\begin{figure}[!ht]
\centering
\begin{subfigure}[b]{0.32\textwidth}
\centering
\includegraphics[trim={0 -3.8cm 0 0},clip,width=\textwidth]{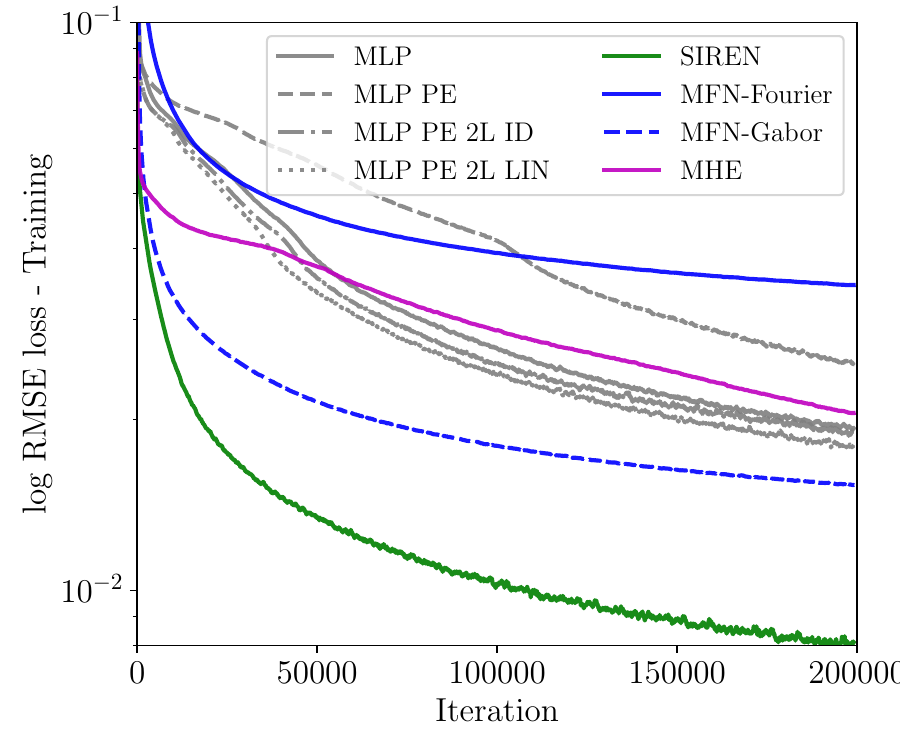}
\caption{Loss profiles for XYZT Aorta.}
\label{fig:xyzt_aorta_profile_nrf}
\end{subfigure}
\hfill
\begin{subfigure}[b]{0.32\textwidth} 
\centering
\includegraphics[width=\textwidth]{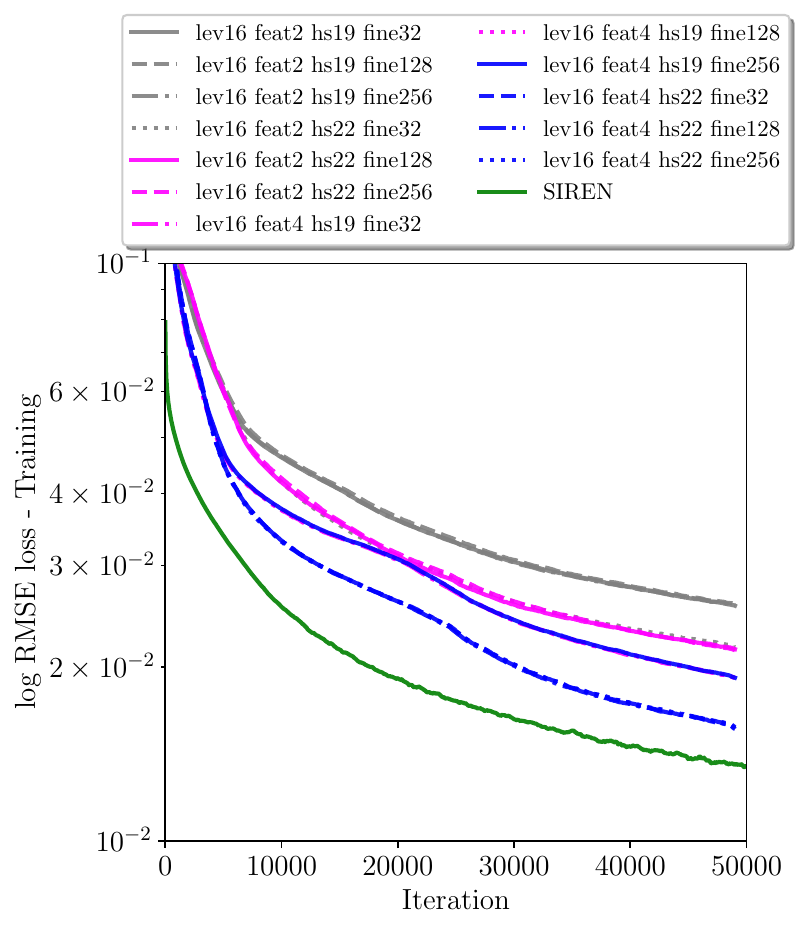}
\caption{Loss profiles for XYZT Aorta from hyperparameter sweep and Cutlass-based implementation in \emph{Tiny-CUDA-nn} library.}
\label{fig:xyzt_aorta_profile_tinycudann}
\end{subfigure}    
\begin{subfigure}[b]{0.32\textwidth} 
\centering
\includegraphics[width=\textwidth]{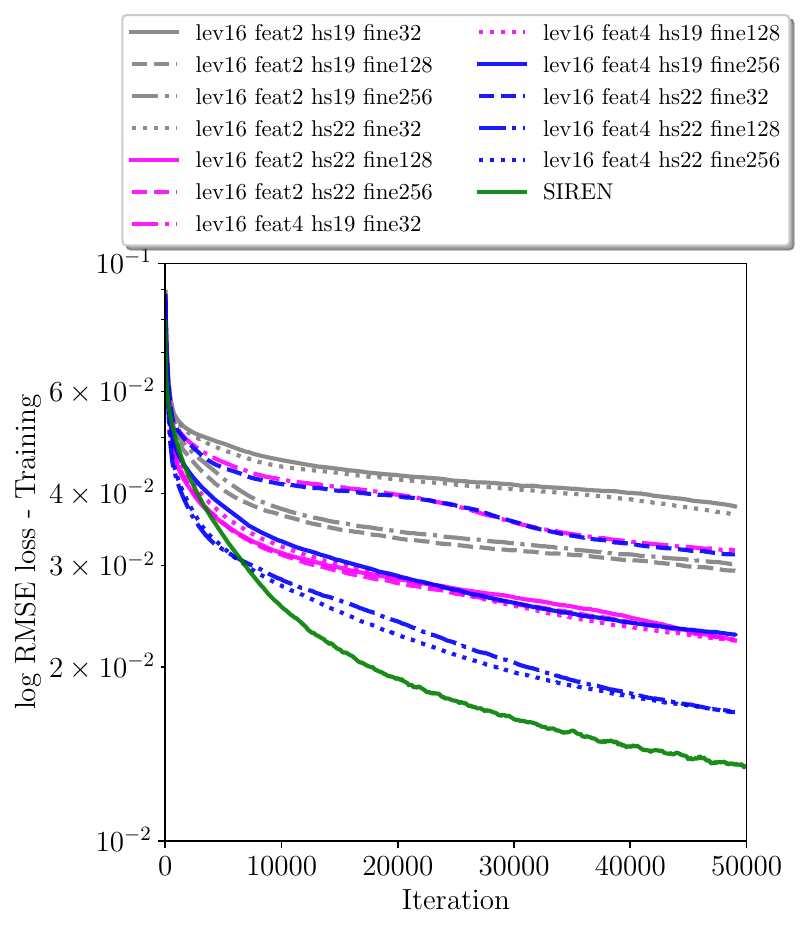}
\caption{Loss profiles for XYZT Aorta from hyperparameter sweep and implementation using the \emph{PhysicsNeMo} library.}
\label{fig:xyzt_aorta_profile_modulus}
\end{subfigure}    
\caption{Loss profiles for different networks and different MHE hyperparameters. Loss profiles are post-processed using a moving average with window size of 1000.}
\label{fig:xyzt_aorta_loss_profiles}
\end{figure}

In Figure~\ref{fig:xyzt_aorta_profiles_space} and~\ref{fig:xyzt_aorta_profile_time} we show the results in terms of space and time profiles generated by a model trained for 200K iterations with 4096 batch size. Resulting errors for the pressure are within 1 mmHg, whereas velocity fluctuations can go up to 5-10 cm/s at diastole, with maximum systolic velocities in the hundreds of cm/s.
Even for this case, sublinear convergence can be observed as a result of the trends in Table~\ref{tab:xyzt_aorta_error_it10K} and~\ref{tab:xyzt_aorta_error_it200K}.
Figure~\ref{fig:xyzt_aorta_profile_nrf} also shows the loss profiles obtained from the various networks smoothed using moving average with window size of 1000. 
Lowest losses are obtained by SIREN followed by MFN-Gabor.
We also tested the losses produced by MHE under various hyperparameter choices for both the implementation provided by the \emph{PhysicsNemo} library and the optimized \emph{tiny-cuda-nn} implementation. The latter contains an efficient \emph{fully-fused} half-precision MLP network with MHE. However, since all the networks we used have 5 layers with 512 neurons (for fully fused the maximum allowed network width is 128), we used an implementation based on the \emph{CUDA Templates for Linear Algebra Subroutines} (CutlassMLP), also provided by the library. 
Resulting loss profiles are shown in Figure~\ref{fig:xyzt_aorta_profile_tinycudann} and~\ref{fig:xyzt_aorta_profile_modulus}, also considering multiple hyperparameter combinations including a number resolution levels $\{16\}$, number of features $\{2,4\}$, logarithm of the hash table size $\{19,22\}$, fine resolution size $\{32,128,256\}$. 
Best results are obtained for 16 resolution levels, 4 trainable features per hash map entry, logarithmic size of the hash map 22, and fine resolution of 128 or 256.
The figures also confirm that SIREN remains the network configuration achieving the best accuracy.

A fully pulsatile simulated hemodynamic field with 4 float components, and a finite element mesh with 554,474 nodes and 180 time steps requires 554,474$\cdot$180$\cdot$4$\cdot$4/1024$^{3}\sim$1.5GB of memory. This leads to considerable compression ratios ranging from $\sim21$ to $\sim350$, where a fully pulsatile simulation can be stored in just a few MBs.

\begin{table}[!ht]
\caption{RMSE results for XYZT Aorta.}
\begin{subtable}[c]{0.45\textwidth}
    \centering
    \caption{Pressure/velocity RMSE for 10K/1024 iteration/batch size hyperparameters - XYZT Aorta.}
    \label{tab:xyzt_aorta_error_it10K}
    \begin{tabular}{l c c}
        \toprule
        {\bf Model} & {\bf RMSE} $p$ & {\bf RMSE} $\Vert\bm{v}\Vert_{2}$\\
        \midrule
        MLP           & 3.853$\times$10$^{-1}$ & 1.365$\times$10$^{+1}$\\
        MLP PE        & 4.559$\times$10$^{-1}$ & 1.414$\times$10$^{+1}$\\
        MLP PE 2L ID  & 2.646$\times$10$^{-1}$ & 1.347$\times$10$^{+1}$\\
        MLP PE 2L LIN & 2.675$\times$10$^{-1}$ & 1.339$\times$10$^{+1}$\\
        SIREN         & 3.118$\times$10$^{-1}$ & $\boldsymbol{6.762}$\\
        MFN-Fourier   & 1.699 & 1.281$\times$10$^{+1}$\\
        MFN-Gabor     & 4.338$\times$10$^{-1}$ & 7.698\\
        MHE           & $\boldsymbol{2.132\times10^{-1}}$ & 8.519\\
        \bottomrule
    \end{tabular}
    
\end{subtable}\hfill
\begin{subtable}[c]{0.45\textwidth}
    \centering
    \caption{Pressure/velocity RMSE for 200K/4096 iteration/batch size hyperparameters - XYZT Aorta.}
    \label{tab:xyzt_aorta_error_it200K}
    \begin{tabular}{l c c}
        \toprule
        {\bf Model} & {\bf RMSE} $p$ & {\bf RMSE} $\Vert\bm{v}\Vert_{2}$\\
        \midrule
        MLP           & 1.326$\times$10$^{-1}$ & 3.260\\
        MLP PE        & 1.529$\times$10$^{-1}$ & 4.749\\
        MLP PE 2L ID  & 1.195$\times$10$^{-1}$ & 3.320\\
        MLP PE 2L LIN & 1.507$\times$10$^{-1}$ & 3.007\\
        SIREN         & $\boldsymbol{7.797\times10^{-2}}$ & $\boldsymbol{1.428}$\\
        MFN-Fourier   & 6.898$\times$10$^{-1}$ & 5.417\\
        MFN-Gabor     & 1.434$\times$10$^{-1}$ & 2.645\\
        MHE           & 1.226$\times$10$^{-1}$ & 3.508\\
        \bottomrule
    \end{tabular}
\end{subtable}
\end{table}

\begin{table}[!ht]
    \centering
    \caption{Pressure/velocity RMSE from MFN-Gabor - XYZT Aorta. RMSE are shown for varying number of iterations and batch size.}
    \begin{tabular}{l c c | l c c}
        \toprule
        {\bf It./Batch size} & {\bf RMSE} $p$ & {\bf RMSE} $\Vert\bm{v}\Vert_{2}$ & {\bf It./Batch size} & {\bf RMSE} $p$ & {\bf RMSE} $\Vert\bm{v}\Vert_{2}$\\
        \midrule
        10K/1024 & 3.931$\times$10$^{-1}$ & 7.405 & 50K/1024 & 2.347$\times$10$^{-1}$ & 4.620\\
        10K/2048 & 4.032$\times$10$^{-1}$ & 6.901 & 50K/2048 & 2.498$\times$10$^{-1}$ & 4.443\\
        10K/4096 & 3.720$\times$10$^{-1}$ & 6.062 & 50K/4096 & 1.948$\times$10$^{-1}$ & 3.773\\
        \midrule
        20K/1024 & 3.258$\times$10$^{-1}$ & 6.198 & 100K/1024 & 1.918$\times$10$^{-1}$ & 3.920\\
        20K/2048 & 2.932$\times$10$^{-1}$ & 5.581 & 100K/2048 & 1.855$\times$10$^{-1}$ & 3.445\\
        20K/4096 & 2.881$\times$10$^{-1}$ & 5.050 & 100K/4096 & 1.721$\times$10$^{-1}$ & 3.223\\
        \midrule
        30K/1024 & 3.238$\times$10$^{-1}$ & 5.754 & 200K/1024 & 1.659$\times$10$^{-1}$ & 3.279\\
        30K/2048 & 2.633$\times$10$^{-1}$ & 4.998 & 200K/2048 & 1.530$\times$10$^{-1}$ & 3.001\\
        30K/4096 & 2.549$\times$10$^{-1}$ & 4.287 & 200K/4096 & 1.269$\times$10$^{-1}$ & 2.684\\
        \bottomrule
    \end{tabular}
    \label{tab:xyzt_aorta_error}
\end{table}

\subsection{Learning Signed Distance Fields for Cardiovascular Anatomies}\label{sec:sdf}
\begin{figure}[!ht]
\centering
\includegraphics[width=\textwidth]{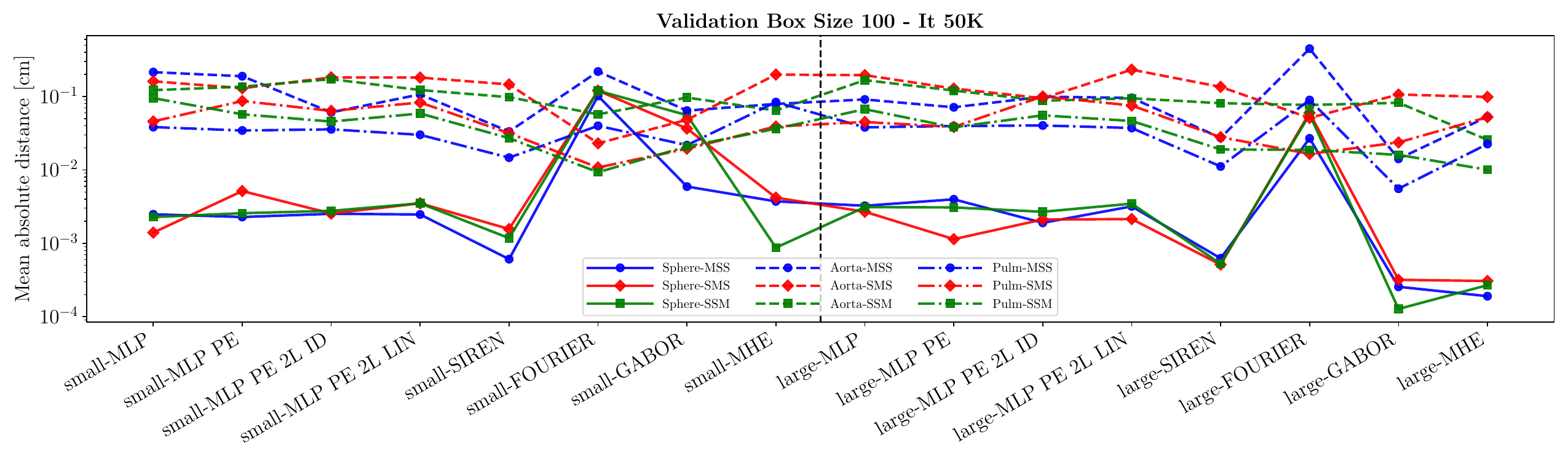}
\includegraphics[width=\textwidth]{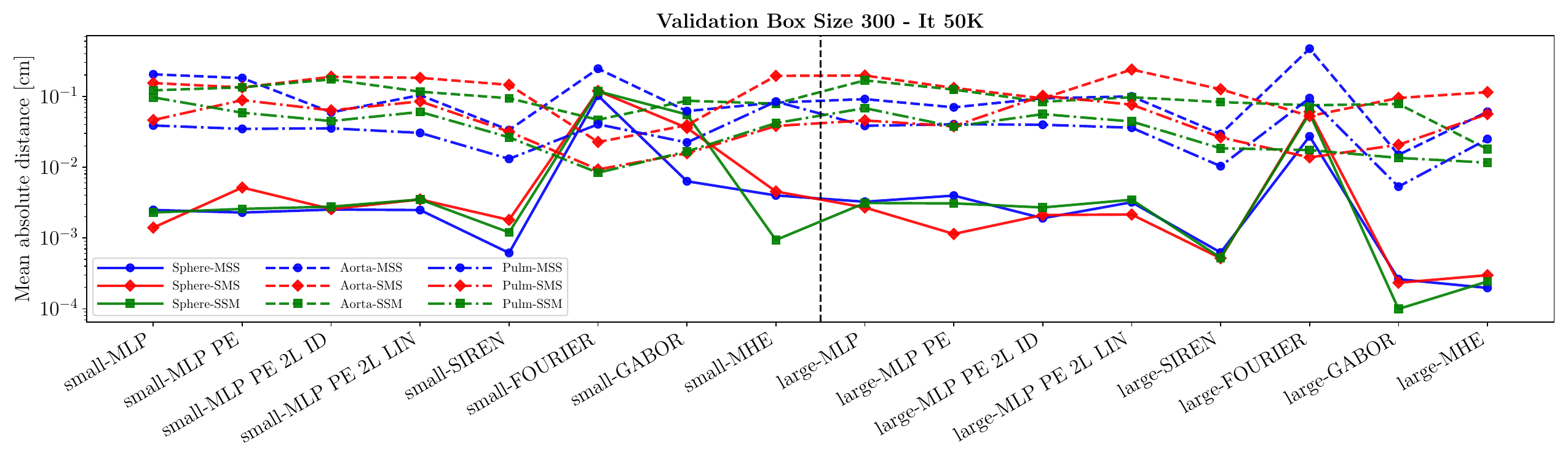}
\caption{Surface reconstructions for all networks, all variations of the training dataset, and $\delta_{\text{SDF}} = 1024$.}
\label{fig:sdf_allnets_1024}
\end{figure}

\begin{figure}[!ht]
\centering
\begin{subfigure}[b]{\textwidth}
\centering
\includegraphics[width=\linewidth]{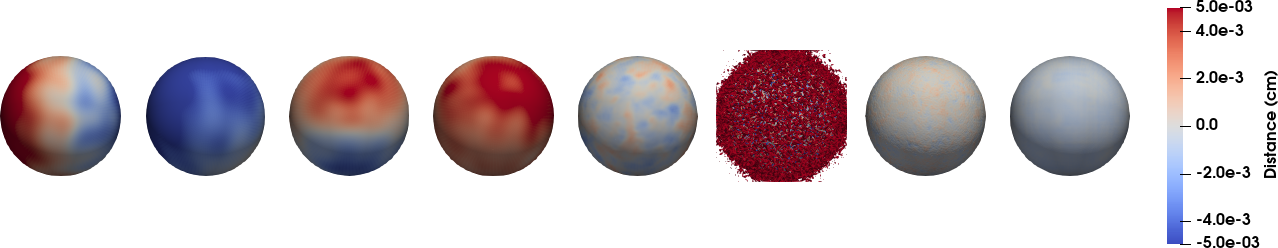}
\caption{Spherical geometry.}
\end{subfigure}

\begin{subfigure}[b]{\textwidth}
\centering
\includegraphics[width=\linewidth]{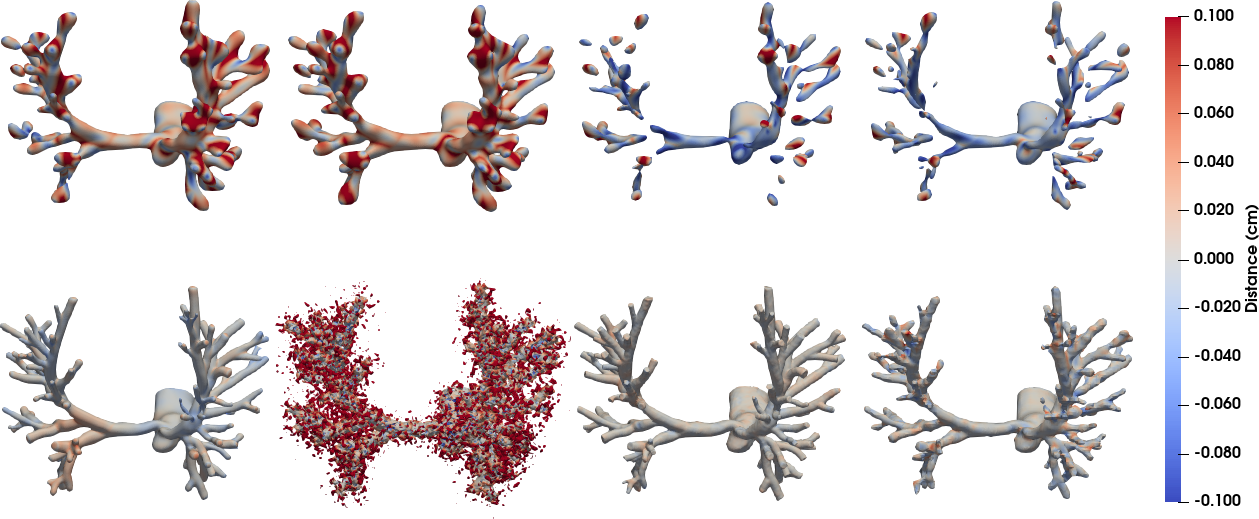}
\caption{Pulmonary anatomy.}
\end{subfigure}

\begin{subfigure}[b]{\textwidth}
\centering
\includegraphics[width=\linewidth]{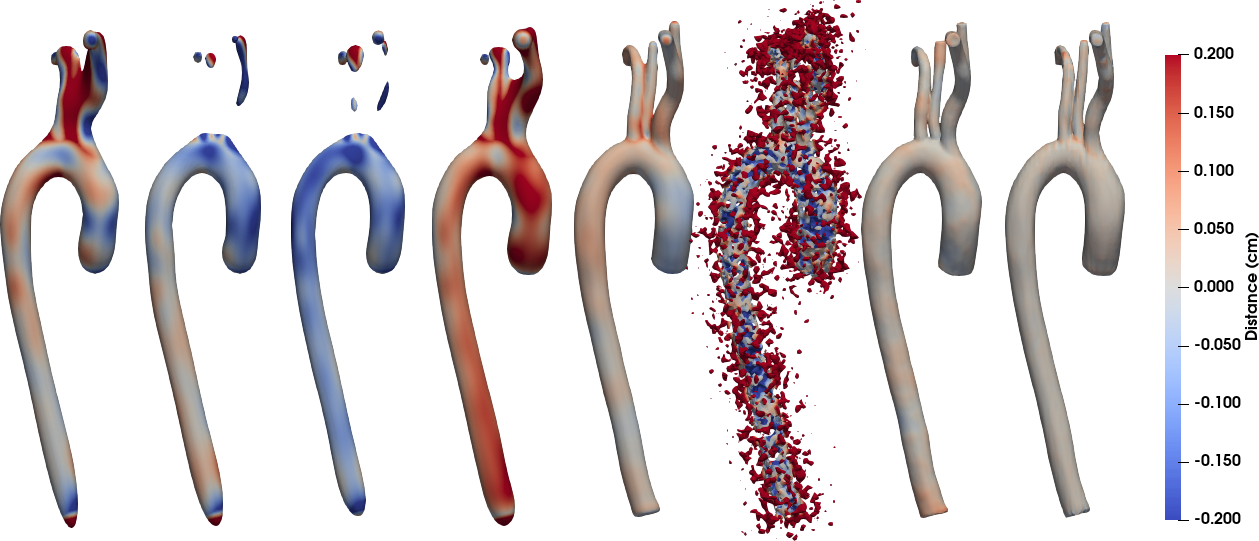}
\caption{Aortic anatomy.}
\end{subfigure}
\caption{Distance between original geometry and INR SDF reconstruction, MSS-large, $\delta_{\text{SDF}} = 1024$, validation grid $100^{3}$. Network from left to right and top to bottom are MLP, MLP PE, MLP PE 2L ID, MLP PE 2L LIN, SIREN, MFN-Fourier, MFN-Gabor and MHE.}\label{fig:sdf_aorta}
\end{figure}

%
We compare the performance of the various networks for the signed distance representation of three geometries including an ideal spherical geometry and two patient-specific anatomies, one pulmonary and one aortic.
The results in term of mean maximum absolute distance are shown in Figure~\ref{fig:sdf_allnets_1024}, showing differences across training datasets (MSS, MSM, SSM and \emph{large} or \emph{small}, see Table~\ref{tab:sdf_large_small}), anatomy and network architecture. 
Due to its isotropic smoothness, the spherical geometry was best approximated by the networks followed by the pulmonary and aortic anatomy, both characterized by $\sim$0.1 cm mean absolute accuracy.
Overall, MHE, NFN-Gabor and SIREN offered the best accuracies, with differences that are evident for the spherical geometry, but not as significant for the pulmonary and aortic anatomy in Figure~\ref{fig:sdf_allnets_1024}. In all these cases MFN-Fourier with default hyperparameters, was found to produce the worst accuracies. However, mean absolute distance might be affected by bias when the zero level set only contains a sub-set of the original geometry. 
Figure~\ref{fig:sdf_allnets_1024} also suggests that \emph{MSS} generally provides the best accuracies over all other ways to generate a training dataset. 
Thus a better understanding of the ability of the network to match fine-scale anatomical features is offered in Figure~\ref{fig:sdf_aorta}. Looking at the pulmonary anatomy and supra-aortic branches in the aortic model, it appears clear how SIREN, MFN-Gabor and MHE have an increasing ability to correctly identify secondary branches. 
Finally, we investigated the effect of perturbing four hyperparameters that are specific of MHE encoding, including the number of grid levels, number of features, logarithm of the hash map size, and size of the fine resolution grid. Best results are obtained for a maximum grid resolution of 32, and increasing the resolution to 256 leads to excessive fine scale variability, unless the number of levels, features or hash map size is also increased. This dependence on hyperparameters makes the MFN-Gabor network generally preferable to MHE, as well as SIREN, even though differences in the selection of the $\omega_{0}$ parameter might affect the quality of the reconstruction. 

\begin{figure}[!ht]
\centering
\includegraphics[width=\textwidth]{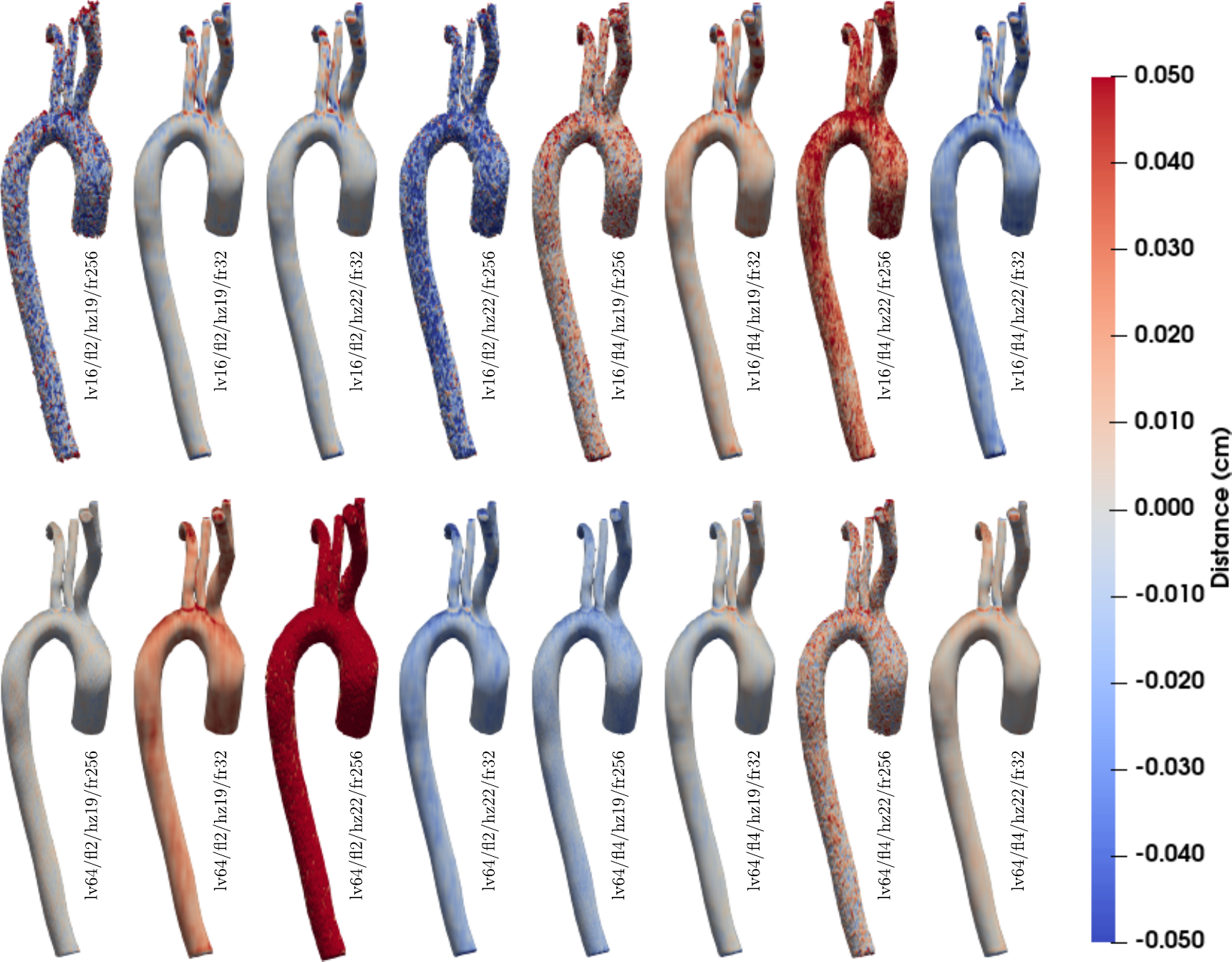}
\caption{Reconstructions of aortic anatomy using MHE positional encoding, varying four hyperparameters, including number of grid levels (\emph{lv}), number of features (\emph{fl}), logarithm of the hash map size (\emph{hz}), and size of the fine resolution grid (\emph{fr}).}
\label{fig:sdf_mhe_variants}
\end{figure}

\subsubsection{Changing the size of the validation grid}\label{sec:sdf_valgrid}

\begin{figure}[!ht]
\centering
\includegraphics[width=\textwidth]{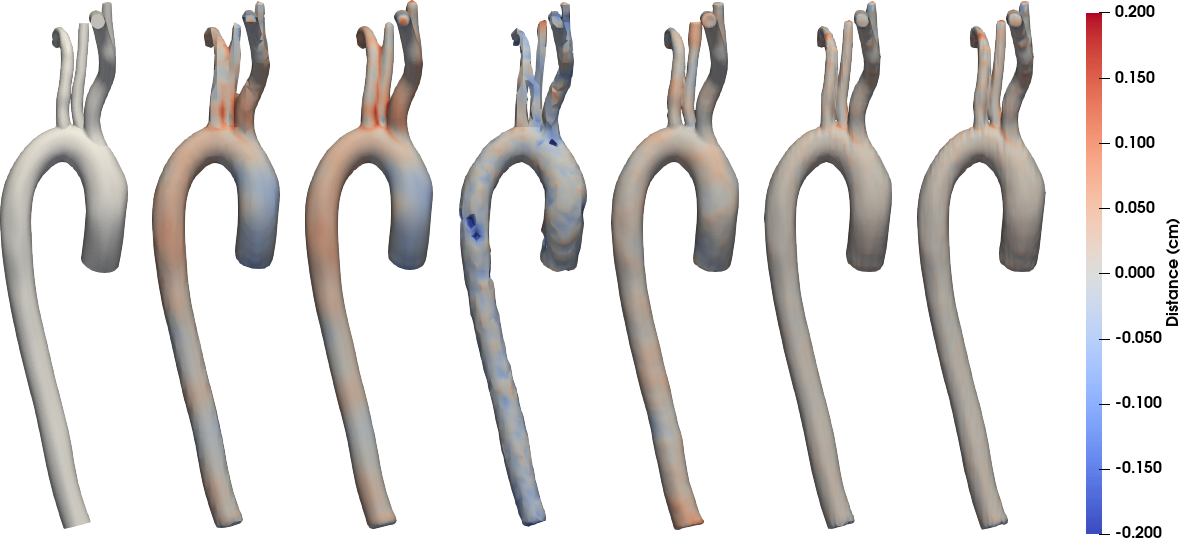}
\caption{Reconstructions with MSS Large. From left to right, reference aortic anatomy, SIREN validation 100$^{3}$, SIREN validation 300$^{3}$, MFN-Gabor validation 100$^{3}$, MFN-Gabor validation 300$^{3}$, MHE validation 100$^{3}$, MHE validation 300$^{3}$.}
\label{fig:sdf_val_grid}
\end{figure}

Surface reconstructions are visualized by first evaluating the signed distance field on a uniform validation grid, then computing the zero levelset in Paraview~\cite{ParaView} using the \emph{contour} filter. In the previous section, we used a 100$^{3}$ validation grid over the unit cube. However, a finer -- or better \emph{adapted} -- validation grid might improve the reconstruction of fine details or sharp edges.
Figure~\ref{fig:sdf_val_grid} shows the effect of increasing the size of the validation grid to 300$^{3}$ for the aortic anatomy with the SIREN, MFN-Gabor and MHE metwork.
Even though the distribution of the absolute distance between the reconstruction and the original geometry is not affected significantly by the size of the validation grid, local reconstruction accuracy might be affected, as seen for the supra-aortic branches in Figure~\ref{fig:sdf_val_grid}.
While these errors are already smaller than the typical precisions resulting from segmenting anatomies from clinical images data, additional improvements can be further explored with adapted, octree-like representations. Future work will further explore these aspects.

\subsection{Signed Distance Representation for Aortic Model Zoo}\label{sec:sdf_zoo}
\begin{figure}[!ht]
    \centering
    \begin{subfigure}[b]{0.9\textwidth}
    \centering
    \includegraphics[width=\textwidth]{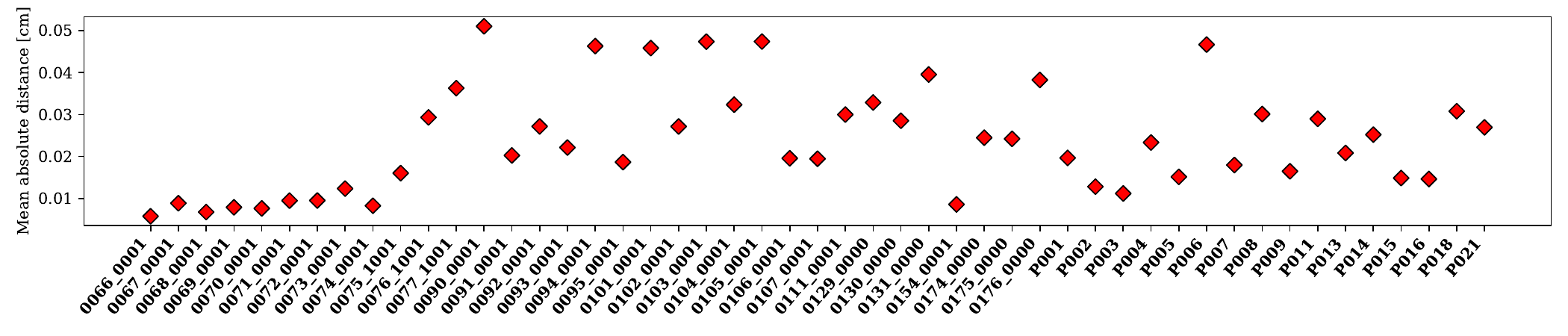}
    \caption{Mean absolute distance from original surface triangulation.}
    \label{fig:zoo_mean_errors}
    \end{subfigure}

    \begin{subfigure}[b]{0.9\textwidth}
    \centering
    \includegraphics[width=\textwidth]{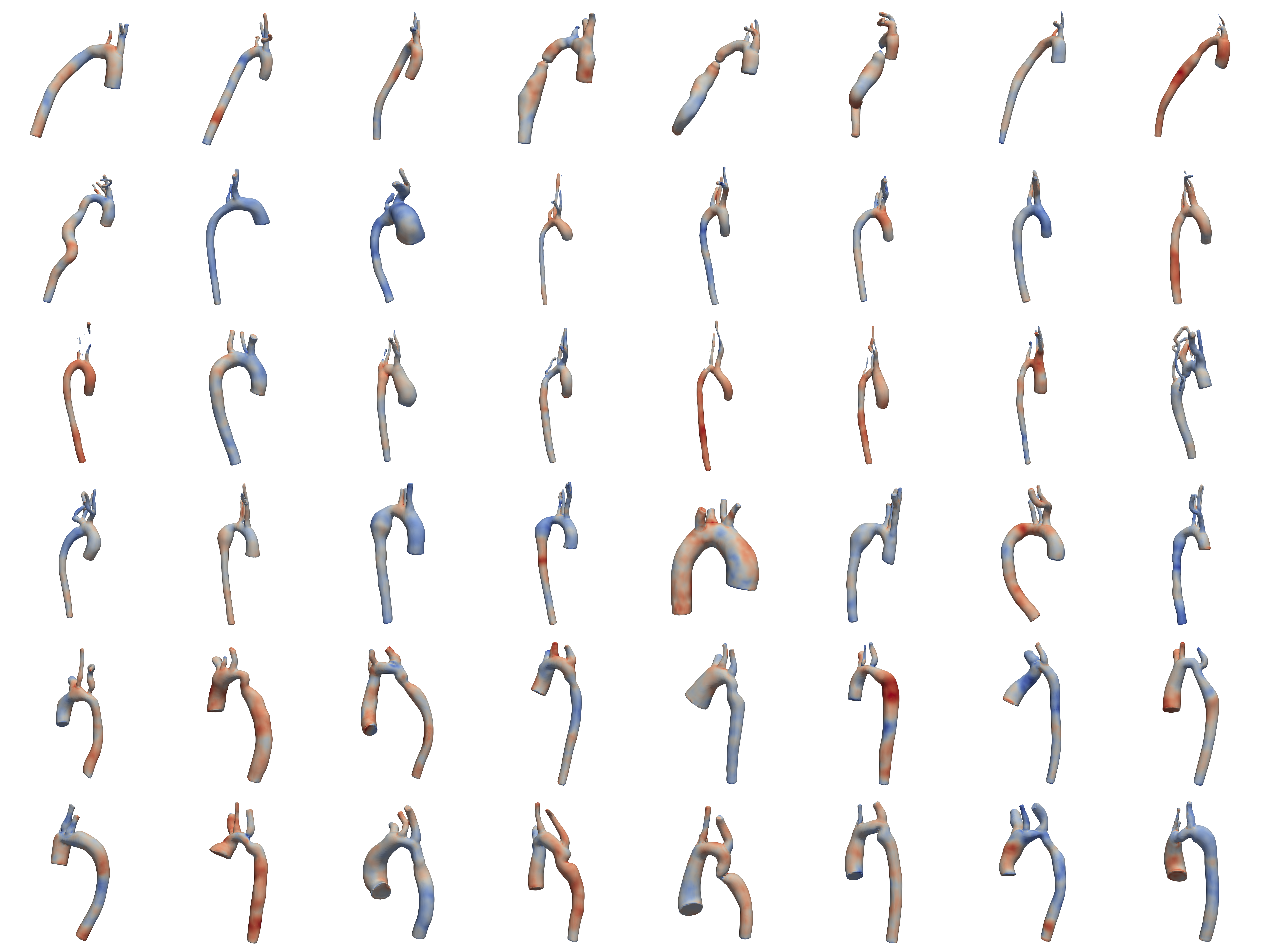}
    \caption{Visualization of model zoo. The surface contour for each model shows the minimum (blue) and maximum distance from the original surface triangulation.}
    \label{fig:zoo_model_errors}
    \end{subfigure}

    \begin{subfigure}[b]{0.9\textwidth}
    \centering
    \includegraphics[width=\textwidth]{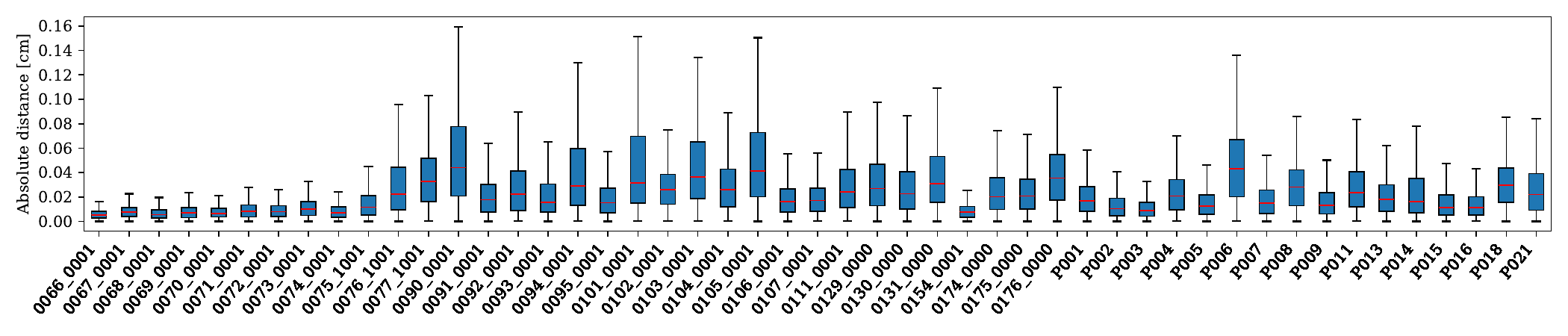}
    \caption{Distribution of absolute distance from original surface triangulation.}
    \label{fig:zoo_error_distributions}
    \end{subfigure}
    \caption{SIREN reconstructions and associated errors for models in the aortic zoo. Results were obtained for a \emph{MSS small} training dataset, 50K iterations and 4096 batch size.}
    \label{fig:model_zoo}
\end{figure}

%
We systematically applied the approach above to a zoo of 48 aortic models from the Vascular Model Repository (see a visualization of the models in Figure~\ref{fig:sdf_zoo_all_models} and the list of model names in Table~\ref{tab:zoo_model_list}).
We reconstructed these anatomies with the three network configurations offering the best performance with the patient-specific anatomies tested in the previous section, i.e., SIREN, MFN-Gabor, and MHE with a \emph{MSS} configuration for the training data.
SIREN resulted in the best performance, with distance errors less than 0.1 cm over all the 48 models, followed by MFN-Gabor, and MHE.
Figures~\ref{fig:zoo_mean_errors} and~\ref{fig:zoo_error_distributions} show the mean error and the entire error distributions for all anatomies, where the maximum error does not exceed 0.16 cm.
Some of the models (e.g., model P011) suffered from the presence of sharp edges, particularly at the caps. Sharp edges at the inlets and outlets might be further distorted as a result of scaling the model so it fits within the unit cube. This leads to errors in computing the signed distance function, to misleading training data, and therefore to poor reconstructions. In such cases, a different approach to scaling the surface anatomy to minimize the presence of sharp edges might solve the issue. 

\section{Discussion and Conclusions}\label{sec:discussion_conclusions}
%
This study investigates the accuracy of neural fields for compression of results fields from numerical hemodynamics.
Our analysis shows that SIREN and MFN-Gabor architectures can consistently deliver accurate pressure and velocity field reconstructions, without the need of laborious hyperparameters tuning.
MHE was found to be competitive but, in some cases, results were found sensitive to the specific hyperparameter selection.
Also, while trainable linear positional encoding was found effective to capture minor but physically relevant features in pulsatile results with dominant time-varying fluctuations, advanced encoding like SIREN, MFN and MHE were found to provide an improved robustness over the entire spectrum of analyzed problems.

When representing complex pressure and velocity fields in space and time from patient-specific simulation, the selected neural representations resulted in pressure errors within 1 mmHg -- for systolic pressures of more than 100 mmHg -- and velocity errors within 10 cm/s observed mainly at diastole -- for velocity fields with maximum systolic velocity norms in the hundreds of cm/s.

The same representations were also able to learn signed distance fields with sub-millimeter accuracy on idealized geometries and patient-specific pulmonary and aortic anatomies.
The best accuracy was achieved using training data consisting in signed distances at spatial locations uniformly sampled over the unit cube. Conversely, samples distributed over the anatomical surface with zero or small perturbation in the orthogonal direction did not seem to have the same effect on the accuracy. 
It should be noted that we considered a loss function where all spatial locations were weighted uniformly.
An increased local density of the validation grid can help to better resolve small geometrical features, better capturing, for example, the separation between small branch vessels. 
Evaluated over a dataset of 48 aortic anatomies from the Vascular Model Repository, average and maximum absolute distances from the original surface models were found always within 0.5 and 1.6 mm distance, respectively.

Future work will focus on the joint representation of fields and geometry, and on representations which also encode uncertainty.
Additionally, we will investigate the ability to perform smooth interpolation without relying on an underlying unstructured mesh.

\section*{Acknowledgments}

This work was partially performed during a sabbatical visit to the CRUNCH research group at Brown University. DES wishes to thank the CRUNCH group and the Division of Applied Mathematics at Brown University for the generous hospitality.
The authors would also like to thank Paris Perdikaris for the discussion on combining trainable linear layers with frequency-based positional encoding.
DES acknowledges support from NSF CAREER award \#1942662, NSF CDS\&E award \#2104831, and NIH grant \#1R01HL167516 {\it Uncertainty aware virtual treatment planning for 
peripheral pulmonary artery stenosis} (PI Alison Marsden). 
DES would like to thank the Center for Research Computing at the University of Notre Dame for providing computational resources and support that were essential to generate the results for this study.

\bibliography{nrf}
\appendix

\section{Appendix}

\subsection{Hyperparameter selection summary}\label{sec:hyper}

\begin{table}[!ht]
\centering
\caption{List of hyperparameters - XT Pipe.}
\begin{tabular}{l c c c}
\toprule
{\bf Parameter} & {\bf All Networks}\\
\midrule
{\bf Batch Size} & 1024\\
{\bf Iteration} & 10K\\
{\bf Learning Rate} & 1.0$\times$10$^{-4}$\\
{\bf Number of Neurons} & 512\\
{\bf Number of Layers} & 5\\
\midrule
{\bf MHE - Levels} & 16 \\
{\bf MHE - Log2 Hash Map Size} & 21 \\
{\bf MHE - Base Resolution} & 2 \\
{\bf MHE - Fine Resolution} & 32 \\
\bottomrule
\end{tabular}
\label{tab:xt_pipe_hyper}
\end{table}

\begin{table}[!ht]
\centering
\caption{List of hyperparameters - XY Zebra.}
\begin{tabular}{l c c c c}
\toprule
{\bf Parameter} & {\bf MLPs} & {\bf SIREN} & {\bf MFN} & {\bf MHE} \\
\midrule
{\bf Batch Size} & 5096 & 5096 & 5096 & 5096 \\
{\bf Iteration} & 20K & 20K & 20K & 20K \\
{\bf Learning Rate} & 1.0$\times$10$^{-4}$ & 1.0$\times$10$^{-4}$ & 1.0$\times$10$^{-4}$ & 1.0$\times$10$^{-4}$\\
{\bf Number of Neurons} & 512 & 512 & 512 & 512 \\
{\bf Number of Layers} & 5 & 5 & 5 & 5 \\
\midrule
{\bf MHE - Levels} & - & - & - & 16 \\
{\bf MHE - Log2 Hash Map Size} & - & - & - & 21 \\
{\bf MHE - Base Resolution} & - & - & - & 2 \\
{\bf MHE - Fine Resolution} & - & - & - & 256 \\
\bottomrule
\end{tabular}
\label{tab:xy_zebra_hyper}
\end{table}

\begin{table}[!ht]
\centering
\caption{List of hyperparameters - XYZ Pipe.}
\begin{tabular}{l c c c c}
\toprule
{\bf Parameter} & {\bf MLPs}  & {\bf SIREN} & {\bf MFN} & {\bf MHE} \\
\midrule
{\bf Batch Size} & 1024 & 1024 & 1024 & 1024 \\
{\bf Iteration} & 10K & 10K & 10K & 10K \\
{\bf Learning Rate} & 1.0$\times$10$^{-4}$ & 1.0$\times$10$^{-4}$ & 1.0$\times$10$^{-4}$ & 1.0$\times$10$^{-4}$\\
{\bf Number of Neurons} & 512 & 512 & 512 & 512 \\
{\bf Number of Layers} & 5 & 5 & 5 & 5 \\
\midrule
{\bf MHE - Levels} & - & - & - & 16 \\
{\bf MHE - Log2 Hash Map Size} & - & - & - & 22 \\
{\bf MHE - Base Resolution} & - & - & - & 2 \\
{\bf MHE - Fine Resolution} & - & - & - & 512 \\
\bottomrule
\end{tabular}
\label{tab:xyz_pipe_hyper}
\end{table}

\begin{table}[!ht]
\centering
\caption{List of hyperparameters - XYZ Aorta.}
\begin{tabular}{l c c c c}
\toprule
{\bf Parameter} & {\bf MLPs} & {\bf SIREN} & {\bf MFN} & {\bf MHE} \\
\midrule
{\bf Batch Size} & 1024 & 1024 & 1024 & 1024 \\
{\bf Iteration} & 10K & 10K & 10K & 10K \\
{\bf Learning Rate} & 1.0$\times$10$^{-4}$ & 1.0$\times$10$^{-4}$ & 1.0$\times$10$^{-4}$ & 1.0$\times$10$^{-4}$\\
{\bf Number of Neurons} & 512 & 512 & 512 & 512 \\
{\bf Number of Layers} & 5 & 5 & 5 & 5 \\
\midrule
{\bf MHE - Levels} & - & - & - & 16 \\
{\bf MHE - Log2 Hash Map Size} & - & - & - & 19 \\
{\bf MHE - Base Resolution} & - & - & - & 2 \\
{\bf MHE - Fine Resolution} & - & - & - & 16 \\
\bottomrule
\end{tabular}
\label{tab:xyz_aorta_hyper}
\end{table}

\begin{table}[!ht]
\centering
\caption{List of hyperparameters - XYZT Pipe.}
\begin{tabular}{l c c c c c}
\toprule
{\bf Parameter} & {\bf MLPs} & {\bf Siren} & {\bf MFN} & {\bf MHE} \\
\midrule
{\bf Batch Size} & 1024 & 1024 & 1024 & 1024 \\
{\bf Iteration} & 10K & 10K & 10K & 10K \\
{\bf Learning Rate} & 1.0$\times$10$^{-4}$ & 1.0$\times$10$^{-4}$ & 1.0$\times$10$^{-4}$ & 1.0$\times$10$^{-4}$\\
{\bf Number of Neurons} & 512 & 512 & 512 & 512 \\
{\bf Number of Layers} & 5 & 5 & 5 & 5 \\
\midrule
{\bf MHE - Levels} & - & - & - & 16 \\
{\bf MHE - Log2 Hash Map Size} & - & - & - & 19 \\
{\bf MHE - Base Resolution} & - & - & - & 2 \\
{\bf MHE - Fine Resolution} & - & - & - & 32 \\
\bottomrule
\end{tabular}
\label{tab:xyzt_pipe_hyper}
\end{table}

\begin{table}[!ht]
\centering
\caption{List of hyperparameters - XYZT Aorta.}
\begin{tabular}{l c c c c c}
\toprule
{\bf Parameter} & {\bf MLPs} & {\bf Siren} & {\bf MFN} & {\bf MHE} \\
\midrule
{\bf Batch Size} & 1024 & 1024 & 1024 & 1024 \\
{\bf Iteration} & 10K & 10K & 10K & 10K \\
{\bf Learning Rate} & 1.0$\times$10$^{-4}$ & 1.0$\times$10$^{-4}$ & 1.0$\times$10$^{-4}$ & 1.0$\times$10$^{-4}$\\
{\bf Number of Neurons} & 512 & 512 & 512 & 512 \\
{\bf Number of Layers} & 5 & 5 & 5 & 5 \\
\midrule
{\bf MHE - Levels} & - & - & - & 16 \\
{\bf MHE - Log2 Hash Map Size} & - & - & - & 19 \\
{\bf MHE - Base Resolution} & - & - & - & 2 \\
{\bf MHE - Fine Resolution} & - & - & - & 32 \\
\bottomrule
\end{tabular}
\label{tab:xyzt_aorta_hyper}
\end{table}

\begin{table}[!ht]
\centering
\caption{List of hyperparameters - SDF.}
\begin{tabular}{l c c c c c}
\toprule
{\bf Parameter} & {\bf MLPs} & {\bf SIREN} & {\bf MFN} & {\bf MHE} \\
\midrule
{\bf Batch Size} & 1024 & 1024 & 1024 & 1024 \\
{\bf Iteration} & 10K & 10K & 10K & 10K \\
{\bf Learning Rate} & 1.0$\times$10$^{-4}$ & 1.0$\times$10$^{-4}$ & 1.0$\times$10$^{-4}$ & 1.0$\times$10$^{-4}$\\
{\bf Number of Neurons} & 512 & 512 & 512 & 512 \\
{\bf Number of Layers} & 5 & 5 & 5 & 5 \\
\midrule
{\bf MHE - Levels} & - & - & - & 16 (64) \\
{\bf MHE - Log2 Hash Map Size} & - & - & - & 19 (22) \\
{\bf MHE - Base Resolution} & - & - & - & 2 (4) \\
{\bf MHE - Fine Resolution} & - & - & - & 32 (256) \\
\bottomrule
\end{tabular}
\label{tab:sdf_hyper}
\end{table}

\begin{table}[!ht]
\centering
\caption{List of hyperparameters - Aortic Zoo.}
\begin{tabular}{l c c c c c}
\toprule
{\bf Parameter} & {\bf MLPs} & {\bf Siren} & {\bf MFN} & {\bf MHE} \\
\midrule
{\bf Batch Size} & 1024 & 1024 & 1024 & 1024 \\
{\bf Iteration} & 10K & 10K & 10K & 10K \\
{\bf Learning Rate} & 1.0$\times$10$^{-4}$ & 1.0$\times$10$^{-4}$ & 1.0$\times$10$^{-4}$ & 1.0$\times$10$^{-4}$\\
{\bf Number of Neurons} & 512 & 512 & 512 & 512 \\
{\bf Number of Layers} & 5 & 5 & 5 & 5 \\
\midrule
{\bf MHE - Levels} & - & - & - & 16 \\
{\bf MHE - Log2 Hash Map Size} & - & - & - & 19 \\
{\bf MHE - Base Resolution} & - & - & - & 2 \\
{\bf MHE - Fine Resolution} & - & - & - & 32 \\
\bottomrule
\end{tabular}
\label{tab:zoo_hyper}
\end{table}

\end{document}